\documentclass{article} 
\usepackage{fullpage}
\usepackage[english]{babel} 

\usepackage[babel]{microtype} 

\usepackage[margin={0.85in, 0.85in}]{geometry}

\usepackage{natbib}

\usepackage[utf8]{inputenc} 
\usepackage[T1]{fontenc}    

\usepackage{amsfonts}
\usepackage{authblk}
\usepackage{float}
\usepackage{subcaption}
\usepackage{graphicx}
\graphicspath{{images/}}
\usepackage{amsmath}
\usepackage{multirow}
\usepackage{makecell}
\usepackage{sidecap}
\usepackage{appendix}
\usepackage{bbm}
\usepackage{xcolor}
\DeclareMathOperator*{\argmax}{arg\,max}

\usepackage{hyperref}
\usepackage{cleveref}
\usepackage{commath}
\usepackage{diagbox}
\usepackage{tabularx} 
\usepackage{adjustbox}

\providecommand{\keywords}[1]{\textbf{\textit{Index terms---}} #1}

\title{Differences Between Hard and Noisy-labeled Samples:\\ An Empirical Study}

\author{Mahsa Forouzesh}
\author{Patrick Thiran}

\affil{Information and Network Dynamics Group (INDY) \protect\\
	School of Computer and Communication Sciences (IC) \protect \\
 \'{E}cole Polytechnique F\'{e}d\'{e}rale de Lausanne} 

\begin{document}

\date{}
\maketitle

\begin{abstract}
  Extracting noisy or incorrectly labeled samples from a labeled dataset with hard/difficult samples is an important yet under-explored topic. 
Two general and often independent lines of work exist, one focuses on addressing noisy labels, and another deals with hard samples. However, when both types of data are present, most existing methods treat them equally, which results in a decline in the overall performance of the model.
In this paper, we first design various synthetic datasets with custom hardness and noisiness levels for different samples. Our proposed systematic empirical study enables us to better understand the similarities and more importantly the differences between hard-to-learn samples and incorrectly-labeled samples. These controlled experiments pave the way for the development of methods that distinguish between hard and noisy samples. Through our study, we introduce a simple yet effective metric that filters out noisy-labeled samples while keeping the hard samples. We study various data partitioning methods in the presence of label noise and observe that filtering out noisy samples from hard samples with this proposed metric results in the best datasets as evidenced by the high test accuracy achieved after models are trained on the filtered datasets. We demonstrate this for both our created synthetic datasets and for datasets with real-world label noise. Furthermore, our proposed data partitioning method significantly outperforms other methods when employed within a semi-supervised learning framework\footnote{Code is available at \url{https://github.com/mahf93/Hard-vs-Noisy}.}.

\end{abstract}

\section{Introduction}
Deep neural networks have revolutionized many applications, especially in the field of image classification, mostly due to the availability of large, high-quality labeled datasets \citep{rawat2017deep}. In practice, obtaining such datasets is often challenging, time-consuming, and expensive, thus leading to the inclusion of label noise in the obtained datasets \citep{roh2019survey}. Label noise can arise for various reasons such as the use of cheap label collection alternatives, for instance crowdsourcing, or the obtainment of the label of an image from the accompanying text on the Web \citep{cordeiro2020survey, frenay2013classification, algan2020label, karimi2020deep}. The problem with label noise is that deep neural networks tend to easily memorize these noisy labels, which can negatively impact their generalization performance \citep{zhang2021understanding}. Consequently, an important line of research is to address this issue. 
Many methods propose to mitigate the effects of label noise, including the use of robust loss functions \citep{ma2020normalized, thulasidasan2019combating, wang2019symmetric, patrini2017making} and modifications to the training procedures \citep{yu2019does, zhang2020distilling, jiang2018mentornet, malach2017decoupling}. Another popular approach to deal with label noise is to use a noisy-label detection method \citep{nguyen2019self,li2020dividemix, huang2019o2u, pleiss2020detecting, pleiss2020identifying}, where some of these methods may require an additional clean validation set for hyper-parameter selection \citep{paul2021deep}.
Noisy-label detection methods can involve data cleansing, where the noisy data is entirely removed from the data set altogether, data re-weighting, where noisy data are given lower weights during training, or re-labeling, where the noisy data is re-annotated by experts. Alternatively, the detected noisy-labeled data could be used as unlabeled data in a semi-supervised learning fashion.

When using noisy-label detection methods, an issue that often arises is their inability to differentiate between noisy-labeled samples and hard-to-learn samples. Hard-to-learn samples, also simply known as hard samples, refer to samples in a dataset that are particularly challenging for the classifier to learn \citep{arpit2017closer, kishida2019empirical, wu2018blockdrop}. Empirical observations have revealed that noisy samples and hard samples share certain characteristics, such as high loss or low confidence or being learned later in the training process. Consequently, when noisy label detection methods are employed, which are based on these characteristics, hard samples are also treated as noisy samples and are either filtered out or given lower emphasis during training. Unfortunately, discarding hard samples may result in gaps in the classifier's knowledge of the true decision boundary, making it crucial to preserve as many hard samples as possible and prioritize their leaning \citep{chang2017active, bengio2009curriculum, wang2020pathological}. It is thus vital to propose noisy-label detection methods capable of distinguishing between noisy-labeled samples and hard samples, keeping as many hard samples as possible while removing noisy ones.

Recent studies that propose noisy-label detection methods which are claimed to retain hard samples lack a quantitative evaluation of such claim, as there is no precise measure to quantify sample hardness \citep{liang2022tripartite, bai2021me, zhu2021hard, zhang2022combating}. To overcome this limitation, in this work, we propose synthetic dataset transformations to simulate varying levels of hardness and noisiness. Although synthetic label noise has been previously studied \citep{zhang2021understanding, rolnick2017deep}, our work, to the best of our knowledge, is the first to introduce synthetic hardness levels and to associate each sample of a dataset with a custom hardness level.
\begin{figure*}[t]
	\centering
	\subfloat[Original Dataset]{\includegraphics[width=.225\linewidth]{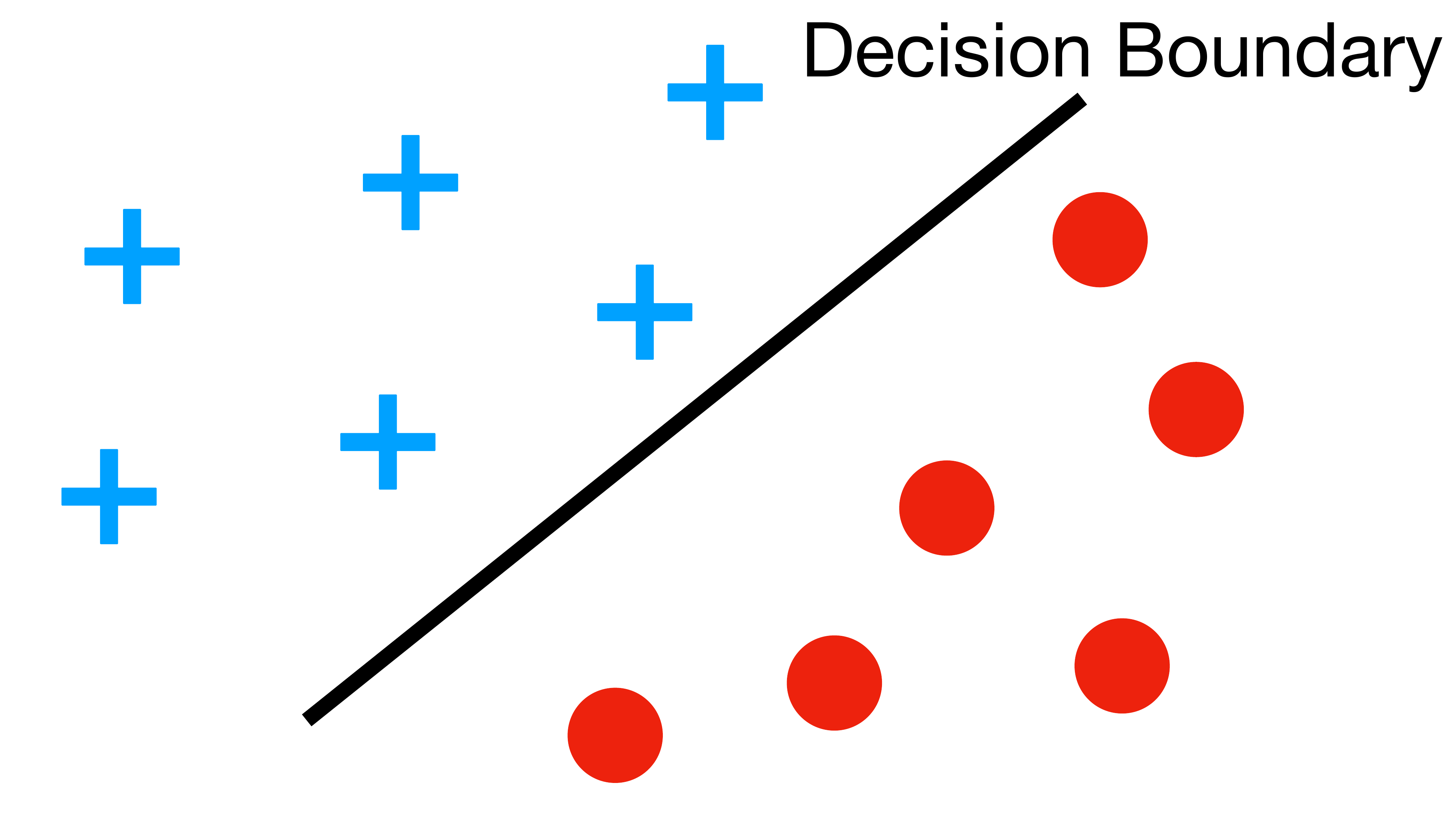}}\quad%
        \subfloat[Imbalance]{\includegraphics[width=.225\linewidth]{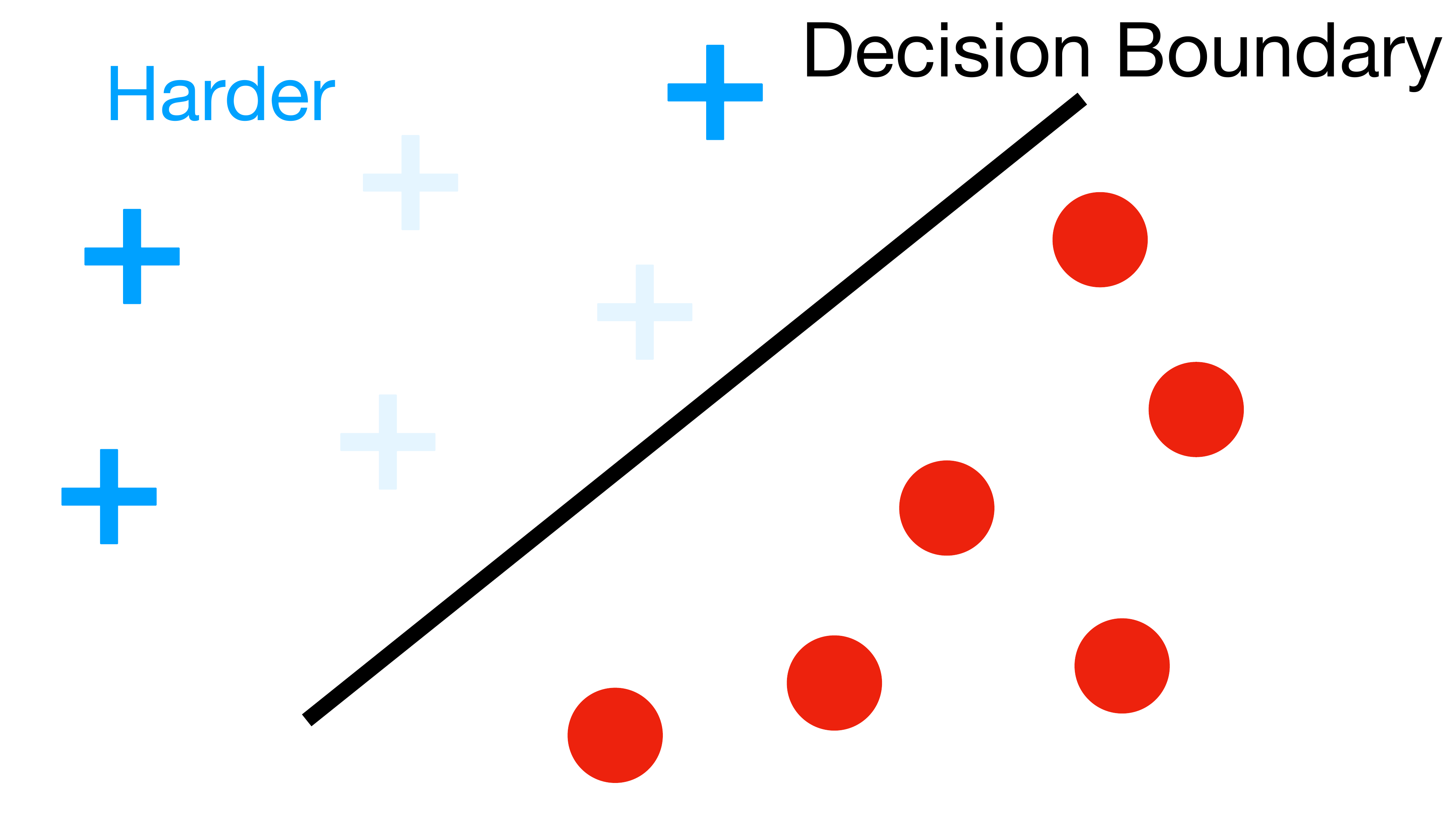}}\quad%
        \subfloat[Diversification]{\includegraphics[width=.225\linewidth]{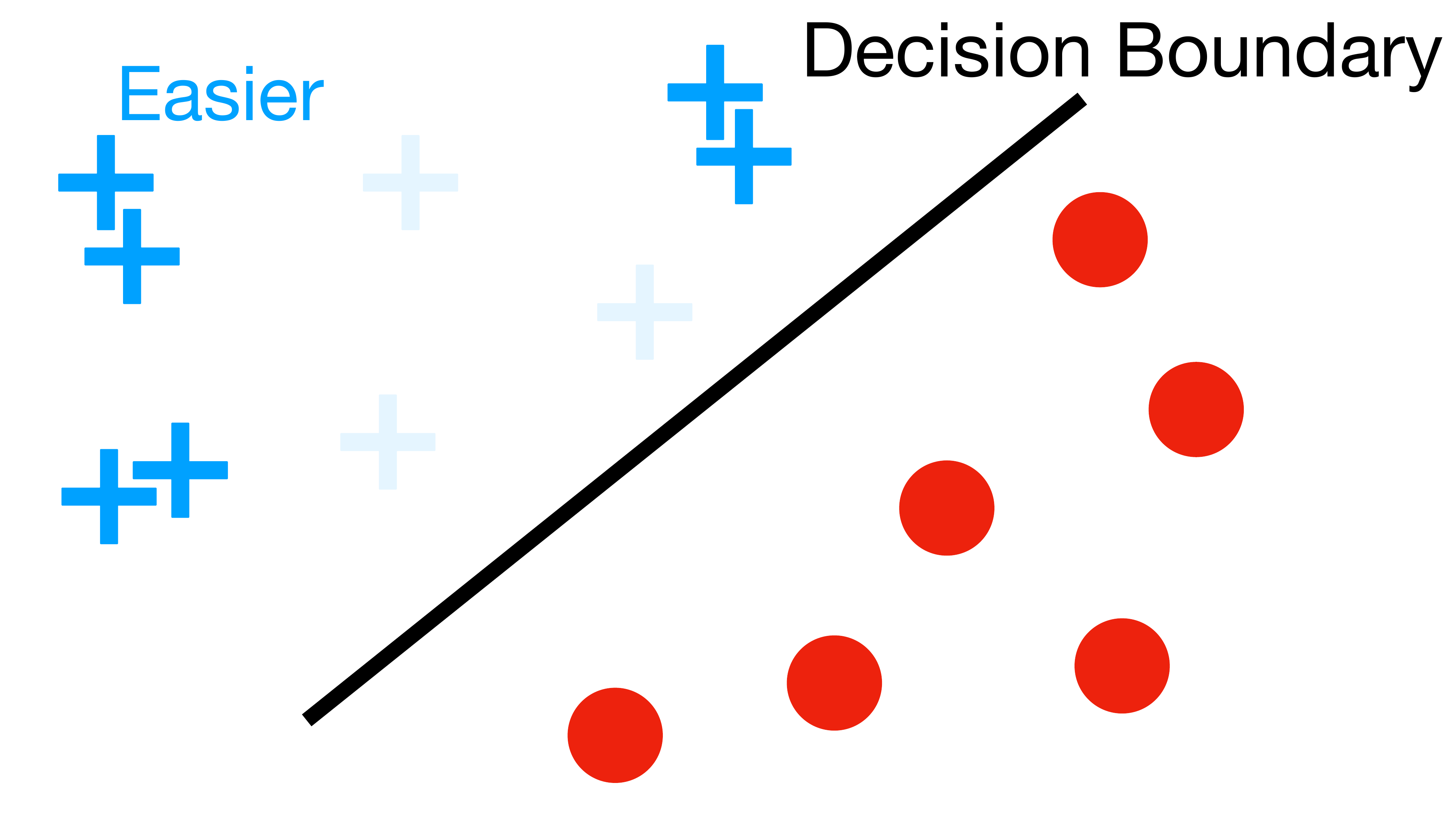}}\quad%
        \subfloat[Closeness to the Decision Boundary]{\includegraphics[width=.225\linewidth]{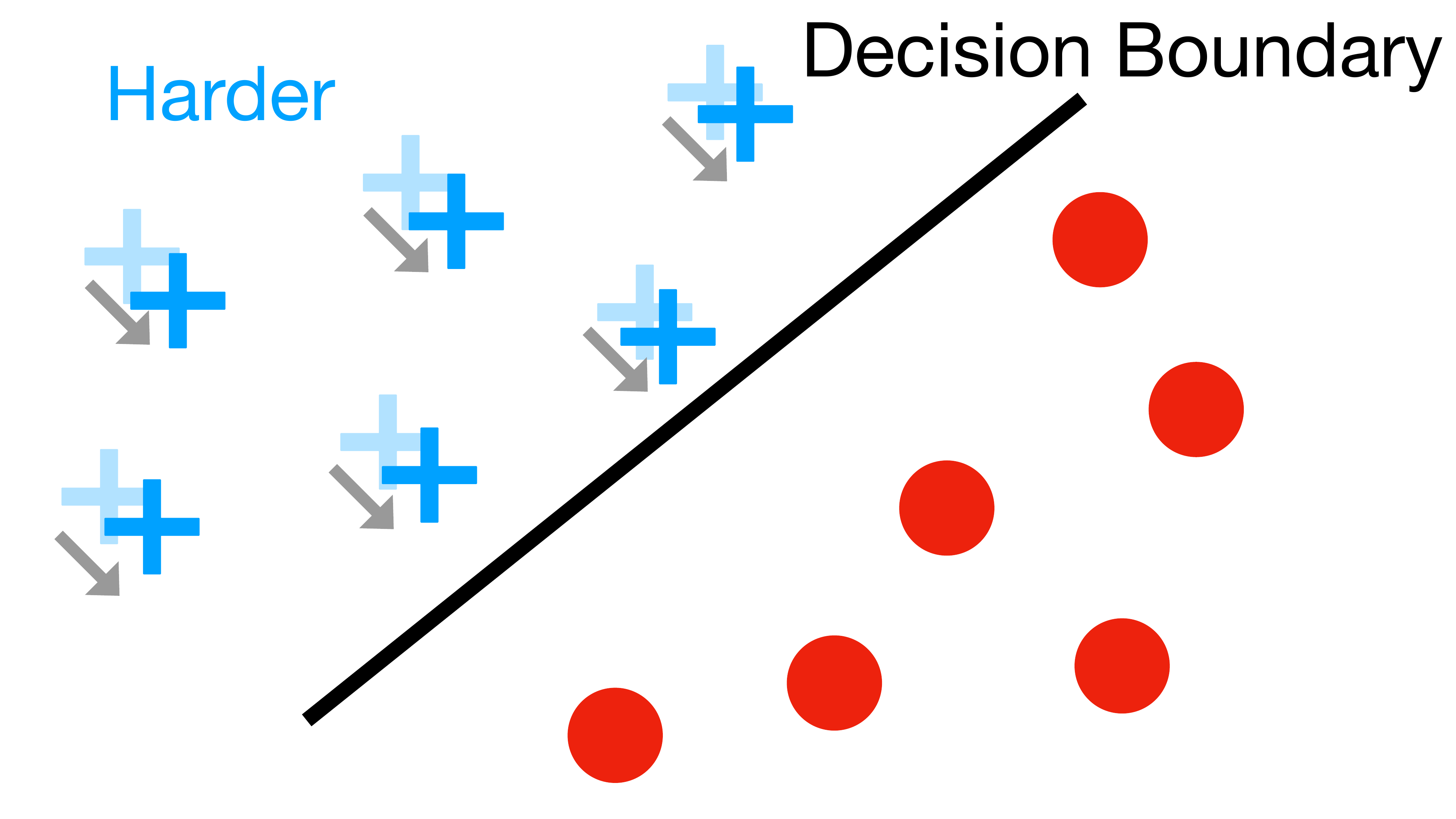}}\quad%
	\caption{A schematic visualization of the hard sample creation approaches. We transform an original dataset (a) into datasets with varying levels of hardness for different classes (b)-(d). We consider three hardness types: imbalance (b), diversification (c), and closeness to the decision boundary (d). In all three transformations, we keep the samples of the red class (circles) unchanged, and transform samples of the blue class (crosses) to become harder (in (b) and (d)) or easier (in (c)), compared to the samples of the red class. In each hardness type, the created dataset has samples with custom hardness levels depending on the degree to which we apply the above transformations.
	}\label{fig:figure1}
\end{figure*}

There can be various reasons that a sample is hard to classify: it can be under-represented in the data set, it can have distinct characteristics from other samples in its class, or it can be close to the decision boundary. To introduce synthetic hardness levels into an original dataset, we propose three main approaches, which involve applying transformations to the original dataset input samples to artificially make them more/less hard to classify.
We refer to these approaches as \emph{hardness types}. We have: (i) imbalance, (ii) diversification, and (iii) closeness to the decision boundary. In the first hardness type (i), sample hardness is introduced by creating an imbalance dataset, and subsampling different classes with varying cardinality. This results in under-represented classes being harder to learn. In the second hardness type (ii), hardness is introduced by making different classes more or less distinct in their samples. 
This is achieved by applying a varying number of data augmentations to different classes, thus resulting in classes with fewer distinct samples and with more augmented samples per distinct sample being easier to learn. In the third hardness type (iii), hardness is introduced by modifying the input samples to be closer to the decision boundary. The decision boundary is estimated using a pre-trained model with a high test accuracy, and the samples are modified to be closer to this estimated decision boundary. A schematic visualization of these different hardness types and transformations from the original dataset is given in Figure~\ref{fig:figure1}. We do not claim these approaches cover all contexts in which hard samples arise, and in practice, a sample might be hard because of any combination of the aforementioned reasons, or some other reason. Yet, these common-sense approaches of introducing sample hardness enable us to perform controlled experiments to assess different metrics/methods in terms of their ability to distinguish between hard and noisy samples.

Our key observation is that the feature embeddings of hard samples become closer to each other during training, whereas noisy-labeled samples do not necessarily exhibit this behavior because of the visual dissimilarity to other samples in the same class. This observation leads us to propose the distance between the feature layer vector of each sample and the centroid feature vector of its assigned class, which we call \emph{static centroid distance} (SCD), as a metric to distinguish between hard and noisy-labeled samples. Next, we propose a label noise detection method based on SCD. While other methods perform well in only one of the two tasks among filtering out noisy samples and retaining hard samples, our method is the only one that performs well in \emph{both} tasks, and consistently so for all hardness types as well as datasets with real-world label noise.
We demonstrate the superior performance of this method for noisy label detection when used for data cleansing as well as semi-supervised learning. Our main contributions are as follows:
\begin{itemize}
	\item We propose a novel approach for synthetically transforming samples to attain custom hardness levels across three hardness types: imbalance, diversification, and closeness to the decision boundary. To the best of our knowledge, we are the first to use controlled experiments to quantitatively assess how different label noise detection methods perform in terms of retaining hard samples.
	\item We study various metrics for detecting noisy labels and show that static centroid distance (SCD) is the most effective metric in distinguishing between hard and noisy-labeled samples. While other metrics remain monotonic by increasing either as a function of hardness or noisiness, we show that SCD is the only metric that is increasing with noisiness but is \emph{not} increasing with hardness.
	\item We propose and evaluate different methods for data cleansing and sample selection and we show that a two-dimensional Gaussian mixture model, which uses the accuracy over the training and SCD as features, performs the best in terms of filtering out noisy samples while retaining hard ones. 
	\item We empirically show that our method produces the best generalization performance when models are trained on the filtered datasets. This holds both in the synthetic datasets and even better in datasets with real-world label noise.  Moreover, if after the data filtration, semi-supervised learning is applied, our method significantly outperforms other label noise detection methods.
\end{itemize}

\section{Background and Related Work}\label{sec:back}
In this section, we first introduce the problem setup developed in this work to produce datasets with custom noisiness and hardness levels. Next, we recall some metrics from previous works that are relevant to our study, and, we introduce static centroid distance (SCD). Finally, we present various partitioning approaches that can be used in conjunction with each metric for sample selection.

We assume that the training set $S$ can be partitioned as $S = S_n \cup S_e \cup S_h$, where $S_n$ are the incorrectly-labeled samples (or noisy samples), $S_e$ are the correctly-labeled and easy-to-learn samples, and $S_h$ are the correctly-labeled and hard-to-learn samples. In practical applications, it is often challenging to clearly distinguish between easy and hard samples in a given dataset, and hence correctly place samples in $S_e$ and $S_h$.
This can be a limitation when studying the performance of different methods in terms of their ability to remove incorrectly labeled samples while retaining hard samples. To address this issue, we synthetically apply the transformation $\mathcal{T}$ to some original dataset $S_{\text{org}}$, such that a spectrum of samples with hardness levels $h \in \{0,1,2,3,4\}$ and noisiness levels $n \in \{0,1,2,3,4\}$ emerge: samples are harder to learn as $h$ increases, and have noisier labels as $n$ increases. We will now discuss how $\mathcal{T}$ transforms $S_{\text{org}}$, with no label noise and with uniform hardness levels among its samples, to a dataset $S = \mathcal{T}(S_{\text{org}})$. This transformation provides us with a knob that we can tune to make alterations, allowing for a systematic study of easy, hard, and noisy samples while offering a good comparison base.

Consider a classification task with input $x \in \mathcal{X}$ and ground truth one-hot label vector $y \in \{0, 1\}^K$, where $K$ is the number of classes. 
The original set with $N_{\text{org}}$ input-output pairs $S_{\text{org}} = \{(x_i, {y}_i)\}_{i=1, \cdots, N_{\text{org}}}$ is transformed into the given training set with $N$ training samples, $S = \mathcal{T}(S_{\text{org}}) = \{(\tilde{x}_i, \tilde{y}_i)\}_{i=1, \cdots, N}$.

\textbf{Noisiness transformation} Let $S^\prime = \{(x_i, {y}_i)\}_{i=1, \cdots, N^\prime} \subset S_{\text{org}}$ be a subset of size $N^\prime$ of the original dataset $S_{\text{org}}$. Transformation $\mathcal{T}_n$ maps $S^\prime$ to $S^\prime_t = \mathcal{T}_n (S^\prime) = \{(x_i, \tilde{y}_i)\}_{i=1, \cdots, N^\prime}$, where with probability $1-q(n)$, $\tilde{y}_i = y_i$, and with probability $q(n)$, the non-zero element of $\tilde{y}_i$ is set uniformly at random at index $j \sim U(1, 2, \cdots, K)$.
The label noise level $q(n) \in [0,1]$ is adjusted as a function of the desired noisiness level $n \in \{0,1,2,3,4\}$. 
Such a dataset transformation is common practice to study label noise in a controlled setting, which is done by fixing some label noise level $q$ for the entire dataset \citep{chatterjee2020coherent}.  In our work, the label noise level $q(n)$ is not fixed for the entire dataset, and depending on the noisiness level $n$, it varies between different subsets of samples.

\textbf{Hardness transformation} Let $S^\prime = \{(x_i, {y}_i)\}_{i=1, \cdots, N^\prime} \subset S_{\text{org}}$, be a subset of size $N^\prime$ from the original dataset $S_{\text{org}}$. We want to transform $S^\prime$ such that the samples of $S^\prime_t = \mathcal{T}_h (S^\prime)$ have hardness level $h \in \{0,1,2,3,4\}$. To the best of our knowledge, such a transformation has not been done in prior work. This transformation is a composition of two transformations, i.e., $\mathcal{T}_h = \mathcal{F}_h \circ \mathcal{I}_h$. The first transformation $\mathcal{I}_h$ is one-to-one, and maps $S^\prime$ into $S^{\prime\prime} = \mathcal{I}_h(S^\prime) = \{(x_j, {y}_j)\}_{j \in J_h}$, where $J_h \subseteq \{1, \cdots, N^\prime\}$. The second transformation $\mathcal{F}_h$ is one-to-many, and maps $S^{\prime\prime}$ into $S^\prime_t = \mathcal{F}_h (S^{\prime\prime}) = \{(\tilde{x}_{i}, {y}_{j}) | i \in I(j) \}_{j \in J_h}$, where $\tilde{x}_{i \in I(j)}$ is a transformed version of input sample $x_j$, and $I(j)$ can be a one-to-many mapping. In Section~\ref{sec:datasets}, we elaborate on the transformation $\mathcal{T}_h$ for each hardness type.

The final transformed dataset with $N$ training samples is $S = \cup_{S^\prime_{h,n}} \mathcal{T}_n \circ \mathcal{T}_h (S^\prime_{h,n})$, where the subsets $S^\prime_{h,n} \subset S_{\text{org}}  \forall h,n \in \{0,1,2,3,4\}$ are determined according to a pre-defined policy. With $S^\prime_{t,h,n} = \mathcal{T}_n \circ \mathcal{T}_h (S^\prime_{h,n})$, and for some hardness threshold $h_{\text{threshold}} \in \{0, 1,2,3,4\}$, the three sets partitioning $S$ are
\begin{align}\label{eq:subsets}
	S_n &= \{(\tilde{x}_i, \tilde{y}_i) \in  S | \tilde{y}_i \neq y_i\},\nonumber \\
    S_h &= \{(\tilde{x}_i, \tilde{y}_i) \in  \cup_{h \geq h_{\text{threshold}}} S^\prime_{t,h,n} | \tilde{y}_i = y_i\}, \nonumber\\	
	S_e &= S \setminus (S_h \cup S_n).
\end{align}
In the dataset partition, we set $h_{\text{threshold}}=4$, hence the samples in $S_h$ consist of the hardest samples in the dataset.

In this work, the classifier trained on the dataset $S$ is a neural network.
For each input sample $\tilde{x}_i$, let $\textbf{p}(\tilde{x}_i) = (p_i^1, p_i^2, \cdots, p_i^K)$ be the prediction probability output vector of the neural network. 
Furthermore, the last layer of the neural network is a fully-connected layer with feature vector $\textbf{h}(\tilde{x}_i) \in \mathbb{R}^m$, where $m$ is the number of units in the feature layer of the neural network. Note that because the parameters of the neural network depend on the epoch during training, 
both vectors $\textbf{p}(\tilde{x}_i)$ and $\textbf{h}(\tilde{x}_i)$ depend on the epoch $t \in \{1, 2, \cdots, T\}$, where $T$ is the maximum number of training epochs. We often remove the explicit dependence on $t$ for simplicity.
Training is done by performing stochastic gradient descent (SGD) on the cross-entropy training loss function at each epoch $t$:
\begin{equation*}
L_{S}(t) = \frac{1}{N} \sum_{i=1}^{N} l_i(t) = -\frac{1}{N} \sum_{i=1}^{N} \sum_{j=1}^{K} \tilde{y}_i^j \log p_i^j(t) = -\frac{1}{N} \sum_{i=1}^{N}  \log p_i^{c_i}(t)
\end{equation*}
where $l_i(t)$ is the training cross-entropy loss of sample $\tilde{x}_i$ at epoch $t \in \{1, 2, \cdots, T\}$. The prediction of the neural network classifier at epoch $t$ for sample $\tilde{x}_i$ is class $\tilde{c}_i = \argmax_j p_i^{j}(t)$, and the confidence of the classifier is the prediction probability $p_i^{\tilde{c}_i}$ of the classifier for the predicted class label. 

\textbf{Metrics}
Here, we recall a few metrics that are introduced in prior work and introduce a new metric, static centroid distance (SCD). In the following sections, we perform a comprehensive study on all these metrics in order to detect which metric, or combination of which metrics, is the best at distinguishing between samples in $S_n$ and $S_h$. In order to remain on a computationally limited budget, and on a practical setting, we study metrics that require only a single model and not an ensemble of models, and that do not require hyper-parameter tuning using a clean validation set. For each sample $\tilde{x}_i$, we compute the following metrics defined below. Throughout our study, we primarily focus on loss, confidence, and SCD metrics. The former two are widely used in the literature due to the valuable information they provide about each sample, while the latter is a metric proposed in our work.

\begin{itemize}
	\item \emph{Loss}: The training loss $l_i(T)$ at the end of the training, i.e., at epoch $T$.
	\item \emph{Confidence}: The prediction probability $p_i^{\tilde{c}_i}(T)$ at the end of the training, i.e., at epoch $T$.
	\item First Prediction Epoch: The epoch $t^*$ such that $\tilde{c}_i(t^*) = c_i(t^*)$ and $\tilde{c}_i(s) \neq c_i(s)$ for $s < t^*$.
	\item Accuracy of Predictions over Training: The accuracy of the classifier predictions for $\tilde{x}_i$ against the assigned class label $c_i$ over the training process, i.e., $\frac{1}{T} \sum_{t=1}^T \mathbbm{1}(c_i(t)=\tilde{c}_i(t))$, where $\mathbbm{1}$ is the indicator function. This is used by \cite{sun2020crssc} to detect noisy samples and to remove them from the dataset. 
	\item AUL: Area under Loss: $\sum_{t=1}^T l_i(t)$, which is used by \cite{pleiss2020detecting} to detect noisy samples.
	\item AUM: Area under Margin: $\frac{1}{T}\sum_{t=1}^T p_i^{\tilde{c}_i}(t)- p_i^{{c}_i^{\prime}}(t)$, where $p_i^{{c}_i^{\prime}}(t)$ is the largest second logit at time $t$. This is used by \cite{pleiss2020identifying} to detect noisy samples.
	\item JSD: Jensen-Shannon divergence between $\textbf{p}_i$ and $\tilde{y}_i$ at the end of the training, i.e., at epoch $T$. Its weighted version, WJSD, is also proposed by \cite{zhang2022combating} to detect noisy samples. 
	\item ACD: Adaptive Centroid Distance, which is the cosine distance between the feature vector $\textbf{h}(\tilde{x}_i)$ and the \emph{adaptive} centroid of feature vectors $\textbf{o}_{c_i}$ at the end of the training, i.e., at epoch $T$.
	The term \emph{adaptive} refers to the computation of $\textbf{o}_{c_i}$ as it is the centroid for samples that the classifier suspects to be in class $c_i$: either these samples have been assigned to another class but the classifier predicts them to be in class $c_i$, or they have been assigned to class $c_i$ and the classifier has high $p_i^{c_i}$. This metric is proposed by \cite{zhang2022combating} to detect noisy samples. 
	\item SCD (in our work)\label{eq:ch5_scd}:  Static Centroid Distance, which is the Euclidean distance between $\textbf{h}(\tilde{x}_i)$ and the \emph{static} centroid of feature vectors $\textbf{o}_{c_i}^s$ at the middle of training, $t^*$, where $\text{training accuracy}(t^*) \geq 50\%$ and $\text{training accuracy}(s) < 50\%$ for $s<t^*$. This is developed in our work as opposed to ACD, as a result of our empirical observations in studying hard samples. The adaptations from ACD to compute SCD are the following: (1) the distance function in SCD is Euclidean distance instead of cosine distance as we are interested in different regions of the feature space and not in different angles/directions of the feature vectors; 
	(2) SCD is computed in the middle of training, where the training accuracy is $50\%$. 
 We observe that in this intermediate stage of training the differences between hard and noisy samples are at their highest; (3) the centroid vector $\textbf{o}_{c_i}^s$ is computed for \emph{all} samples that are assigned to class $c_i$, and not only for those samples which the classifier predicts to be in class $c_i$. Because for hard classes, the classifier, which might not have learned them correctly, does not predict samples that actually belong to class $c_i$, the distance to $\textbf{o}_{c_i}$ becomes rather meaningless. We observe that relying too much on the classifier to compute this centroid, which is done in the computation of ACD, results in inferior performance, especially when the quality of the data is low. In Table~\ref{tab:ablation}, we provide an ablation study that transitions from using different variations of ACD to using SCD, and we can observe that using SCD results in both removing incorrectly labeled samples and retaining correctly labeled hard samples.
\end{itemize}

\textbf{Partitioning Methods}
Using the above metrics, one can then cluster the available samples in $S$ in order to find estimates for $S_e$, $S_h$, and $S_n$, which we refer to by $\tilde{S}_e$, ${\tilde{S}}_h$ and ${\tilde{S}}_n$, respectively. Note that our overall goal is to find the estimated noisy subset ${\tilde{S}}_n$ and the estimated clean subset ${\tilde{S}}_c = {\tilde{S}}_e \cup {\tilde{S}}_h$, but not precisely the two subsets ${\tilde{S}}_e$ and ${\tilde{S}}_h$. 
\begin{itemize}
	\item Thresholding (\emph{Thres} in short) - In this method, a threshold value is chosen for the specific metric, which can be the average value over the samples or any other predetermined value. The samples are then partitioned into two regions, one containing samples whose values on the metric are greater than the threshold, and the other one containing the samples whose values on the metric are smaller than the threshold.
	Because we work with values in the dataset that might have outliers (in particular for hard samples and noisy samples), we choose the median value as the default threshold when using this method.
	\item 1d-GMM - A one-dimensional Gaussian mixture model is used for partitioning the dataset into multiple subsets or partitions by modeling the values of the metric for each sample as a mixture of several Gaussian distributions, and each Gaussian distribution represents one of the partitions. 
	\item 2d-GMM - Similarly to 1d-GMM, a two-dimensional GMM can partition the dataset by using two metrics instead of only one. 
\end{itemize}

\section{Dataset Design with Different Hardness Levels}\label{sec:datasets}
\begin{figure}
 \centering
	\includegraphics[width=0.5\textwidth]{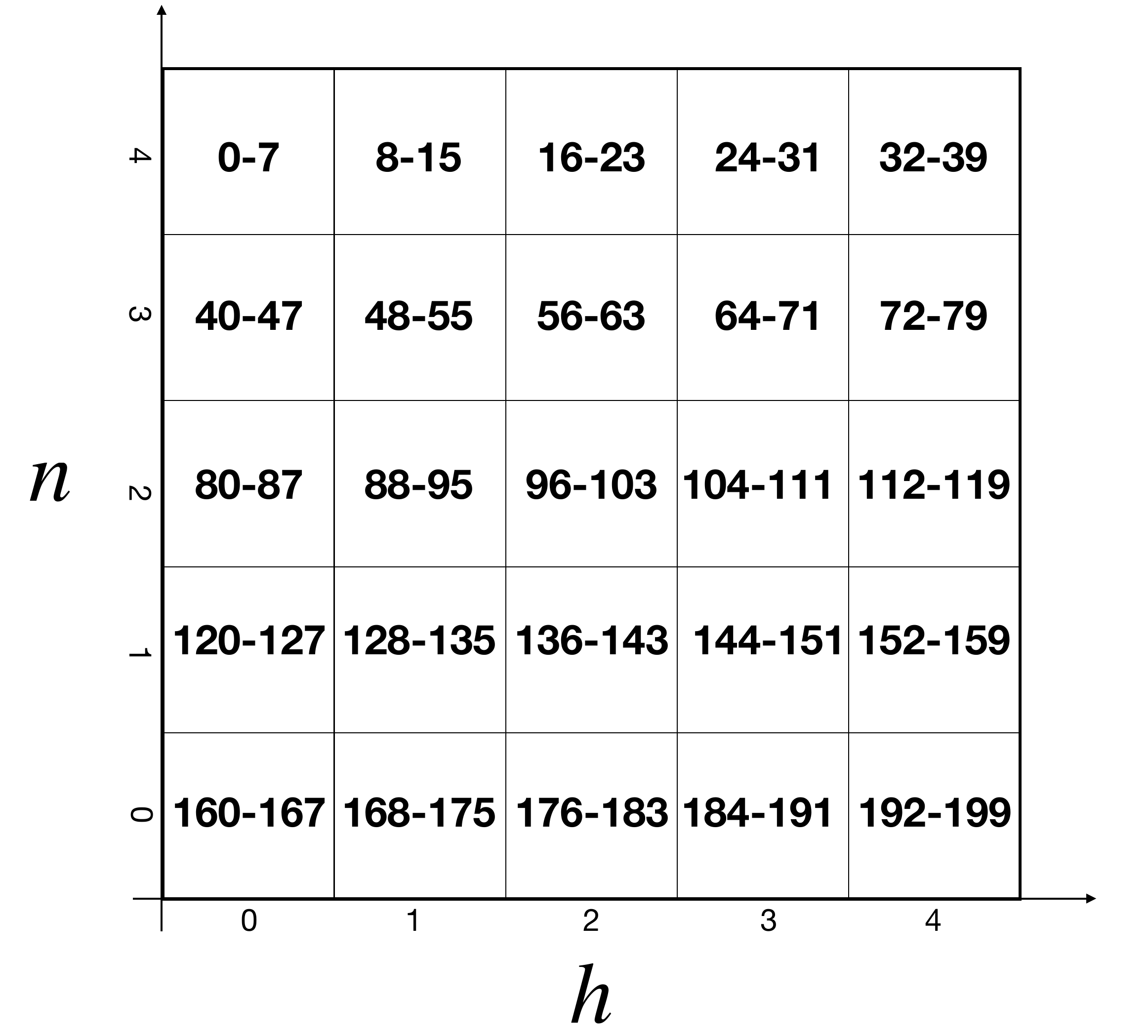}
	\captionsetup{width=0.95\textwidth}
	\caption{For a dataset with $200$ classes, the hardness level $h$ and noisiness level $n$ of samples in the dataset are determined by their class, as shown on this $2$-dimensional figure. For each pair $(h,n)$, we allocate class number $c = 40*(4-n)+8*h+\beta$ where $\beta \in \{0,1,\cdots,7\}$, so that each pair $(h,n)$ is represented by $8$ different classes. For example, samples with $h=0$ and $n=4$ are in classes $0$ to $7$. The pre-defined policy to determine subsets $S^\prime_{h,n} \subset S_{\text{org}}$ is according to this figure.
 }\label{fig:ch5_spectrum}
\end{figure}

In this section, we discuss about the hardness and noisiness transformations $\mathcal{T}_h$ and $\mathcal{T}_n$. Sections~\ref{sec:ch5_im}, \ref{sec:ch5_di}, and \ref{sec:ch5_db} provide details of the transformation $\mathcal{T}_h$ in each of the three hardness types: imbalance, diversification, and closeness to the decision boundary, respectively. Finally, Section~\ref{sec:noi} provides details of $\mathcal{T}_n$.\\
The original dataset $S_{\text{org}}$ is taken here as the TinyImagenet dataset \citep{le2015tiny} with $200$ classes. We have chosen this dataset because it possesses sufficient complexity to capture the underlying input-output relations involved in intricate image classification tasks. Later in this paper, we assess the generalizability of our findings by testing them on real-world datasets that contain noisy labels.\\
To ensure consistency within each class, we assign the same levels $h$ and $n$ to all samples in the same class. To provide a comprehensive spectrum of samples, we vary $h$ and $n$ along two orthogonal axes and across five levels, resulting in a $2$D space with four quadrants (and a total of $25$ different sample categories): easy-clean ($h=0$ and $n=0$), easy-noisy ($h=0$ and $n=4$), hard-clean ($h=4$ and $n=0$), and hard-noisy ($h=4$ and $n=4$). Each sample in $S_{\text{org}}$ is defined by a pair $(h, n)$ of levels according to Figure~\ref{fig:ch5_spectrum}, and the set of samples with levels $(h,n)$ is denoted by $S^\prime_{h,n}$, its size by $N^\prime_{h,n}$. We first apply a hardness transformation (one of the three hardness types) to $S^\prime_{h,n}$ to produce $S^{\prime\prime}_{h,n} = \mathcal{T}_h(S^\prime_{h,n})$. We then apply the noisiness transformation to $S^{\prime\prime}_{h,n}$ and finally get the transformed subset $S^\prime_{t,h,n} = \mathcal{T}_n \circ \mathcal{T}_h (S^\prime_{h,n})$. 
In order to demonstrate the effectiveness of the hardness transformation in terms of actually making samples more or less hard, we use metrics that are widely accepted from the literature, such as loss \citep{kishida2019empirical, loshchilov2015online}, and confidence \citep{swayamdipta2020dataset, chang2017active, wang2020pathological}. Prior studies show that hard samples have larger losses and lower confidence. By examining the relation between loss/confidence and the hardness levels assigned by our approach, we find that our hardness level assignments are well-founded.

\subsection{Hardness via Imbalance}\label{sec:ch5_im}
\begin{figure*}[t]
	\centering
	\subfloat[Hardness via Imbalance\label{fig:ch5_proof_im}]{\includegraphics[width=.22\linewidth]{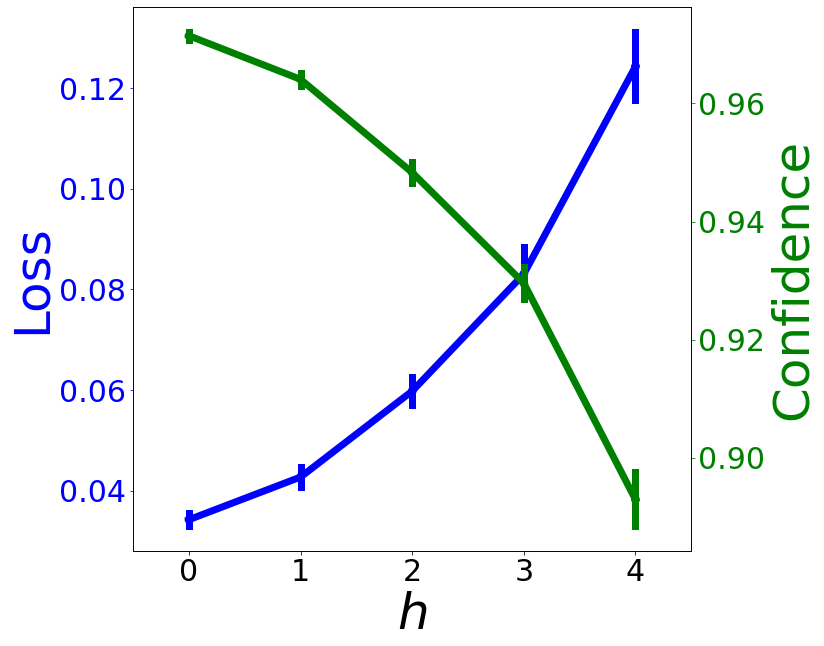}}\quad
	\subfloat[Hardness via Diversification\label{fig:ch5_proof_di}]{\includegraphics[width=.235\linewidth]{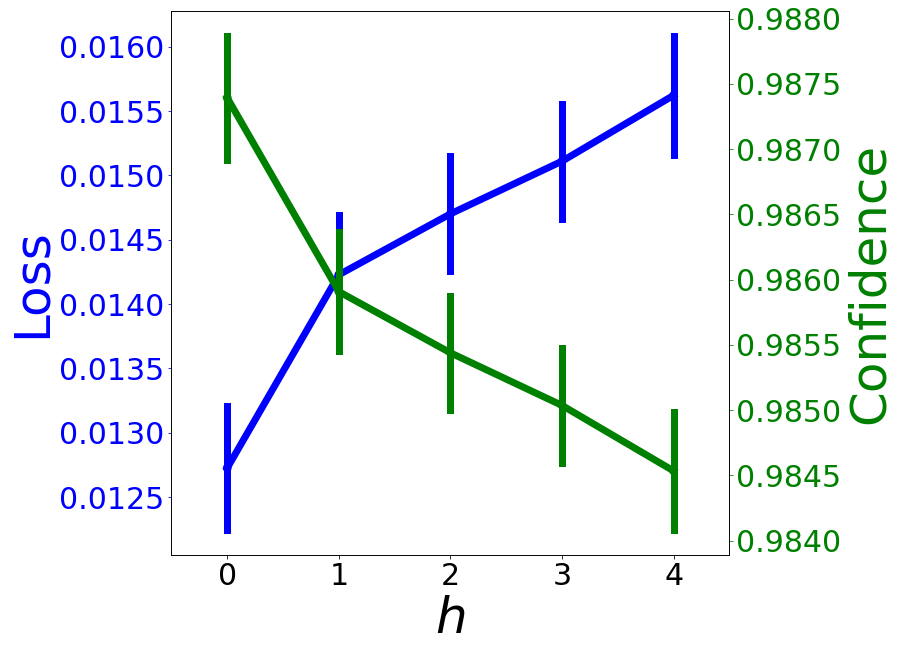}}\quad
	\subfloat[Hardness via Closeness to the Decision Boundary\label{fig:ch5_proof_db}]{\includegraphics[width=.235\linewidth]{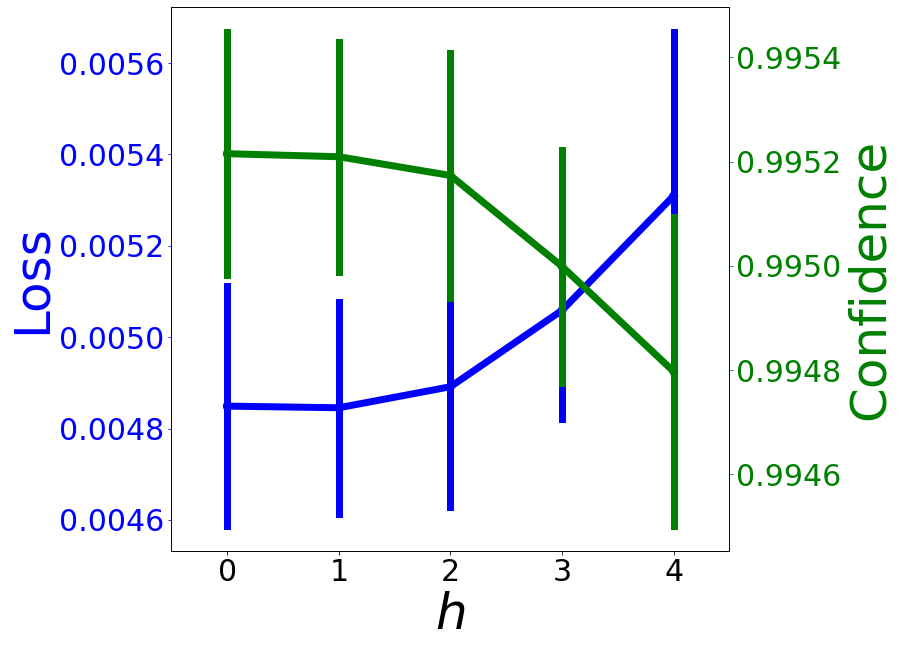}}\quad
 \subfloat[Noisiness\label{fig:ch5_hardness_naive}]{\includegraphics[width=.225\linewidth]{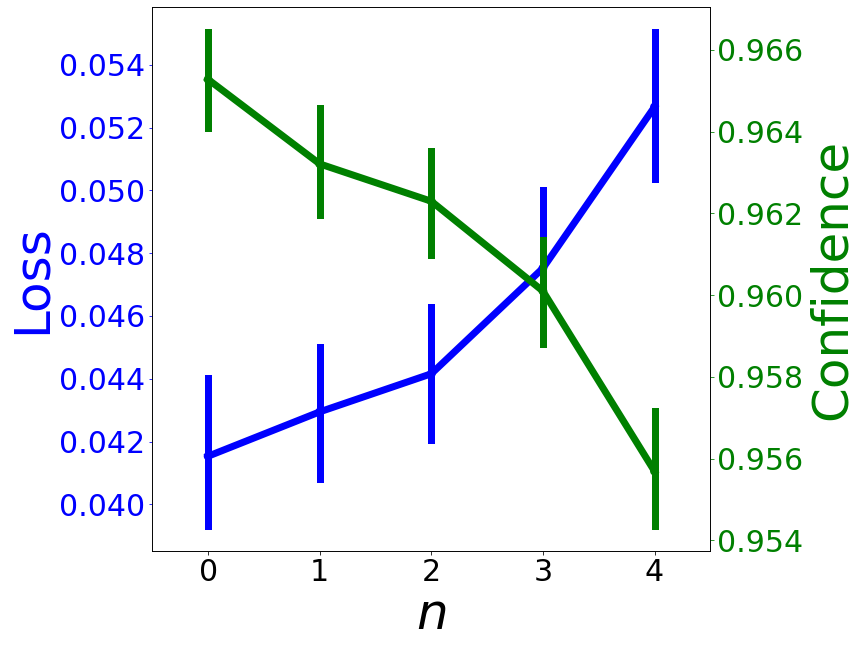}}\quad
	\caption{In each of the three created datasets, we observe that samples in classes with higher~$h$ have generally traits that correspond to their difficulty. These traits include having a higher loss value and having a lower confidence (or prediction probability). 
		To further show the significance of loss and confidence values as $h$ increases, we perform a one-way ANOVA test, which results in the following F-statistics and p-values: \textbf{(a) Hardness via Imbalance:} for the loss we have F=$672$ and p$<1e-100$, for the confidence we have F=$1519$ and p$=0$; 
		\textbf{(b) Hardness via Diversification:} for loss we have F=$234$ and p$<1e-190$, for confidence we have F=$237$ and p$<1e-200$; 
		\textbf{(c) Hardness via Closeness to the Decision Boundary:} For loss we have F=$5.7$ and p$<1e-3$; for confidence we have F=$5.8$ and p$<1e-3$. 
		We can observe that in all three settings, the F-statistics are relatively large and the p-values are small. We conclude that our assigned $h$ values do indeed indicate sample hardness and in all three settings, our approaches for creating hard samples are justified. \textbf{(d) Noisiness:} We observe that, similarly to the effect of hardness increase discussed in (a)-(c), an increase in the level of noisiness increases loss and decreases confidence. Therefore, if we rely solely on these metrics to identify and remove noisy-labeled samples, hard samples can also be mistakenly removed from the dataset. This is problematic because hard samples contain valuable information about the underlying data distribution and should not be ignored.
	}\label{fig:ch5_hardness_proof}
\end{figure*}

In this dataset, we make samples with different levels $h$, by making the dataset imbalanced. 
To achieve this, we subsample the dataset so that the number of samples for each class with hardness $h$ is $X/2^h$, where $X$ is the maximum number of samples per class in the dataset. 
With this approach, the number of samples per class decreases exponentially as $h$ increases, which makes these samples under-represented hence more difficult to learn. We have
\begin{align*}
	S^{\prime\prime}_{h,n} = \mathcal{T}_h (S^\prime_{h,n}) = \mathcal{I}_h (S^\prime_{h,n}) = \{(x_j, y_j)\}_{j \in J_h}
\end{align*}
where $J_h \subseteq \{1, \cdots, N^\prime_{h,n}\}$ uses stratified sampling so that there are $X/2^h$ samples per class. In this hardness type, the transformation $\mathcal{F}_h$ is the identity map and $\mathcal{T}_h = \mathcal{I}_h$.\\
\textbf{Are samples in $S_{h}$ with this hardness type actually harder to learn?} As we can see in Figure~\ref{fig:ch5_proof_im}, samples that are assigned with a larger hardness level $h$ have indeed a higher loss and lower confidence values. We conclude that samples with $h=4$, which are those in $S_h$, are the hardest for the classifier to learn compared to the rest of the samples.

\subsection{Hardness via Diversification}\label{sec:ch5_di}
Compare the following two sets of samples: (i) $16$ images of different cats possibly with different breads, (ii) $16$ images of the same cat with different image augmentations, (augmentations can be rotation, flipping, scaling and blurring \citep{shorten2019survey}). Which of the two sets of samples is harder for a classifier to learn? We can argue that the first set is more difficult, because the classifier requires to learn some common features that belong to $16$ different images instead of $16$ variations of the same image. The first set of samples are more \emph{diverse} and such diversification is the second hardness type we consider. It is also observed by \cite{kishida2019empirical}, that easy examples in the datasets are visually similar to each other whereas hard samples are visually diverse.\\
In this dataset, we keep the number of samples balanced between classes, but we vary the diversity of samples per class. To do so, we maintain the original samples in hard classes in order to preserve their level of diversity/difficulty. For samples in the easy classes, we perform data augmentation techniques, such as rotation or flipping, to make samples less diverse and hence easier to learn. 
By changing the number of data augmentations used in each level of hardness~$h$, this approach enables us to generate a range of hardness levels within a single dataset. In particular, for hardness level $h$, we first subsample the number of samples per class by $X/2^{4-h}$, where $X$ is the maximum number of unique samples per class. Then, we create $2^{4-h}-1$ new augmented samples per sample. As a result, all classes have the same overall number of samples and the dataset is balanced. We have
\begin{align*}
S^{\prime\prime}_{h,n} = \mathcal{T}_h (S^\prime_{h,n}) &= \mathcal{F}_h \circ \mathcal{I}_h (S^\prime_{h,n})
= \mathcal{F}_h \left(\{(x_j, y_j)\}_{j \in J_h}\right)\\
&= \{(\tilde{x}_{i}, y_j)| i \in I_h(j)\}_{j \in J_h},
\end{align*}
where $J_h \subseteq \{1, \cdots, N^\prime_{h,n}\}$ uses stratified sampling so that there are $X/2^{4-h}$ samples per class, $I_h(j) = \{j + a \cdot \abs{J_h}\}_{a \in \{0, \cdots, 2^{4-h}-1\}}$, and for $i \in I_h(j), i\neq j $, $\tilde{x}_{i}$ is an augmented version of $x_j$. \\
\textbf{Are samples in $S_{h}$ with this hardness type actually harder to learn?} As we can see in Figure~\ref{fig:ch5_proof_di}, samples that are assigned with a larger hardness level $h$, have higher loss and lower confidence values. 
Hence, we conclude that samples with $h=4$, which are in $S_h$, are the hardest for the classifier to learn.

\subsection{Hardness via Closeness to the Decision Boundary}\label{sec:ch5_db}
In this dataset, we produce hard samples by modifying them to be closer to the decision boundary. To achieve this, we must first identify the true decision boundary for the classification task at hand. As we are working with the TinyImagenet dataset, the true decision boundary is unknown. The closest estimate to the true decision boundary that we found is a DeiT model \citep{touvron2021training} that was pre-trained on ImageNet and fine-tuned on TinyImagenet, with a training accuracy of $98.8\%$ and a test accuracy of $90.88\%$ on the original train and test sets of TinyImagenet. Then, we create samples that are closer to the decision boundary of this model, hoping that they will also be closer to the decision boundary of the actual ground truth model. \\
We have 
\begin{align*}
S^{\prime\prime}_{h,n} = \mathcal{T}_h (S^\prime_{h,n}) = \mathcal{F}_h (S^\prime_{h,n}) = \{(\tilde{x}_j, y_j)\}_{j \in J_h},
\end{align*}
where $J_h = \{1, \cdots, N^\prime_{h,n}\}$, and $\tilde{x}_j$ is a transformation of the sample $x_j$, such that $\tilde{x}_j$ is closer to the decision boundary compared to $x_j$. In this hardness type the transformation $\mathcal{I}_h$ is an identical map and $\mathcal{T}_h = \mathcal{F}_h$.
Our hard sample creation is done similarly to the fast gradient sign method for creating adversarial samples, which is introduced by \cite{goodfellow2014explaining}. For an input sample $x_i$, this method  creates adversarial input
$$
x_i^{\text{adv}} = x_i + \epsilon \quad \text{sign} \left( \nabla_{x} l_i\right)
$$
where $\epsilon$ is a hyper-parameter used for scaling the added noise. For creating adversarial samples, $\epsilon$ should be large enough to change the label of the sample.
For our dataset creation however, and unlike with adversarial sample creation, we make sure that the resulting sample $\tilde{x}_i$ does \emph{not} change its label, by adjusting $\epsilon$. 
Different levels of $h$ are obtained by adjusting the value of $\epsilon$, and hence $\epsilon(h)$ is a function of the hardness level $h$. Higher values of $\epsilon$ move the sample closer to the decision boundary, making it harder to learn with a new model. After generating new samples, we keep only those that maintain their original class label, as our goal is to increase the difficulty of samples in their original class. It is critical to choose the appropriate range of $\epsilon(h)$, as a value that is too high can cause most samples to change their predicted class and be excluded from the dataset. This would result in the remaining samples being easier to learn instead of harder, because they were originally very easy and far from the decision boundary. Therefore, we carefully choose the range of $\epsilon(h)$ during our dataset creation, which results in a smaller variation of the loss/confidence as a function of the hardness level $h$, as shown in Figure~\ref{fig:ch5_proof_db}. These samples become hard, particularly for models that make decisions near the decision boundary, which is why in this setting, unlike the other settings, we use models which are pre-trained on the ImageNet dataset instead of using randomly initialized models. \\
\textbf{Are samples in $S_{h}$ with this hardness type actually harder to learn?} As shown in Figure~\ref{fig:ch5_proof_db}, samples that are assigned with a larger hardness $h$ value, have higher loss and lower confidence values. The loss values vary however less with the hardness level $h$ than in the two previous hardness types, because the sensitivity of the range of $\epsilon(h)$ must be large enough to bring the sample close to the boundary, but small enough to keep it from changing its label. Figure~\ref{fig:ch5_proof_db} shows that samples with $h=4$, which are in $S_h$, are still the hardest for the classifier to learn.

\subsection{Addition of Label Noise; Its Similarities to Hardness}\label{sec:noi}
After creating each of the hard datasets, we add label noise to samples of the dataset according to Figure~\ref{fig:ch5_spectrum}. 
For each sample with noisiness $n$, we assign the label noise level of $q(n) = \delta * n / 4$, where $\delta \in \left[0,1\right]$ is the maximum label noise level of the created dataset (by default we use $\delta = 0.4$ in our experiments). Label-noise level refers to the probability of the label being replaced uniformly at random with one of the class labels. For example, samples with $n=1$, have $0.25 \delta$ probability of having a noisy label. 
To prevent confusion between samples in classes with different levels of noisiness $n$, we limit the choice of class labels to only those with the same $n$ when assigning noisy labels. This ensures that samples are assigned to a class with $n=0$ if and only if they actually belong to that class. Overall, for the set $S^{\prime\prime}_{h,n} = \{(\tilde{x}_i, y_i)\}_{i \in \{1, \cdots, N^{\prime\prime}_{h,n}\}}$, 
we have
\begin{align*}
S^\prime_{t,h,n} = \mathcal{T}_n (S^{\prime\prime}_{h,n}) = \{(\tilde{x}_i, \tilde{y}_i)\}_{i \in \{1, \cdots, N^{\prime\prime}_{h,n}\}},
\end{align*}
where with probability $1-q(n)$, $\tilde{y}_i = y_i$, and with probability $q(n)$, the non-zero element of $\tilde{y}_i$ is at index $j \sim U(1, 2, \cdots, K)$, where $U$ is the uniform distribution. \\ 
In Figure~\ref{fig:ch5_hardness_naive}, we observe that increasing $n$, similarly to increasing $h$ as discussed in previous sections, results in a higher loss, and a lower confidence.  
We observe that the same correlation sign between loss/confidence and hardness $h$ also exists between loss/confidence and noisiness $n$. This indicates that hardness and noisiness exhibit similar characteristics, and that removing noisy samples from the training dataset by using loss or confidence could also lead to the inadvertent removal of correctly-labeled hard-to-learn samples.
This highlights the need for more advanced metrics/methods to identify and preserve hard samples while removing noisy ones.

\section{Easy-Hard-Noisy Data Partitioning and Training}
In this section, we use the metrics and methods discussed in Section~\ref{sec:back} to partition the synthetic datasets constructed in Section~\ref{sec:datasets} into two subsets: the estimated clean subset $\tilde{S}_c$ and the estimated noisy subset $\tilde{S}_n$. We evaluate the effectiveness of these methods and metrics in this partitioning task, with a focus on the quality of the filtering between hard and noisy samples. We then compare the performance of different partitioning methods based on the test accuracy of models trained on the estimated clean subset produced from each method.
It is important to note that although the training datasets can contain samples with issues, such as hardness or noisiness in certain classes, the test datasets do not contain any such noisy-labeled or imbalanced samples. Therefore, traditional evaluation metrics, such as test accuracy, provide a fair representation of the generalization performance of models in each setting.

\subsection{Partitioning}

We compare different methods that partition the available dataset into two subsets: $\tilde{S}_c$ and $\tilde{S}_n$. Our primary objective is to include the incorrectly labeled samples within $\tilde{S}_n$, which is the focus of previous studies as well, but in addition, we are also interested in including all hard samples within $\tilde{S}_c$. Therefore, we aim to achieve high recall for hard samples, i.e., $\text{Recall}_{\text{h}} = \frac{|{\tilde{S}}_c \cap {{S}}_h|}{|{S}_h|}$, while maintaining a high recall for noisy samples, i.e., $\text{Recall}_{\text{n}} = \frac{|{\tilde{S}}_n \cap {{S}}_n|}{|{S}_n|}$. Although having a high precision for noisy samples, i.e., $\text{Precision}_{\text{n}} = \frac{|{\tilde{S}}_n \cap {{S}}_n|}{|{\tilde{S}}_n|}$, is desirable, it is less critical. Even if some easy samples are mistakenly included in $\tilde{S}_n$, this is not too harmful because they can be either relabeled or not used in training. This is because such easy and correctly labeled samples are often redundant in the training set, and are likely to be inexpensive to label or replace \citep{paul2021deep}.\\
\begin{figure}[t]
	\centering
	\subfloat{\includegraphics[width=.3\linewidth]{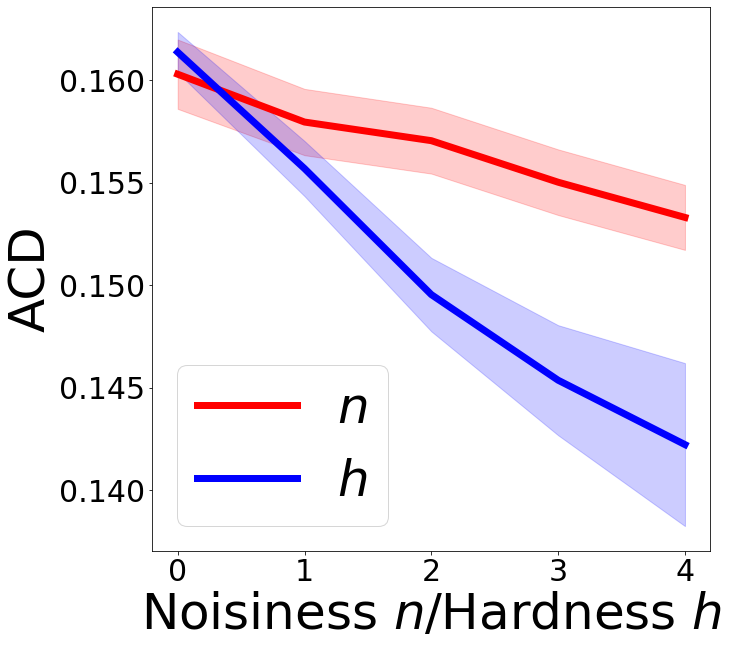}}\quad%
	\subfloat{\includegraphics[width=.3\linewidth]{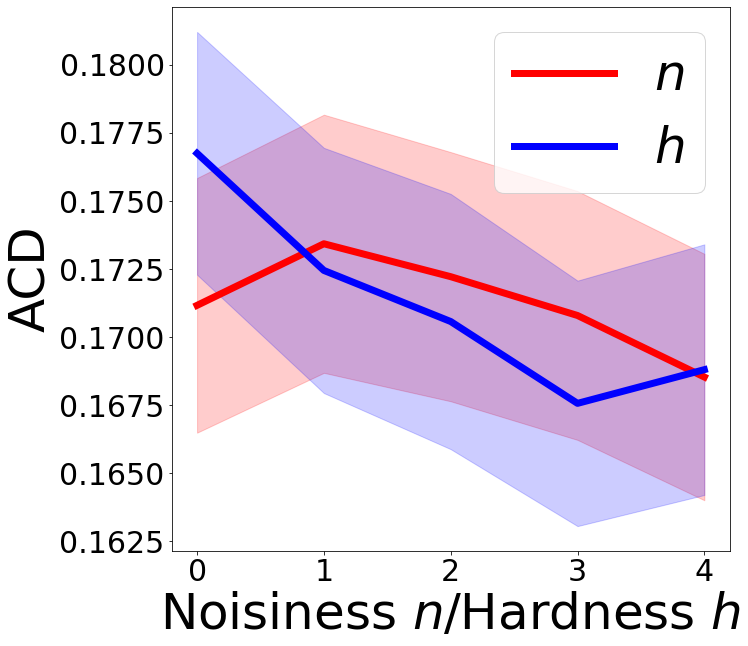}}\quad%
	\subfloat{\includegraphics[width=.3\linewidth]{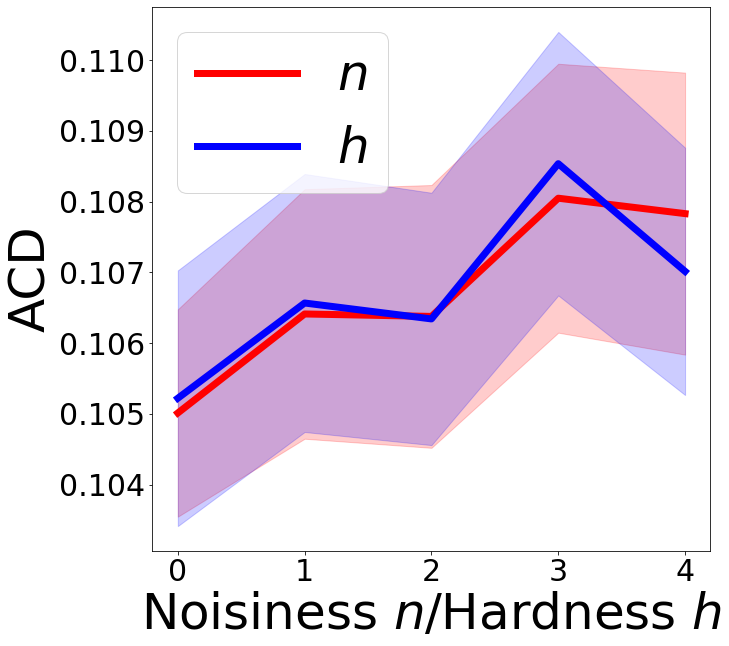}}\\
	\subfloat[Hardness via Imbalance]{\includegraphics[width=.3\linewidth]{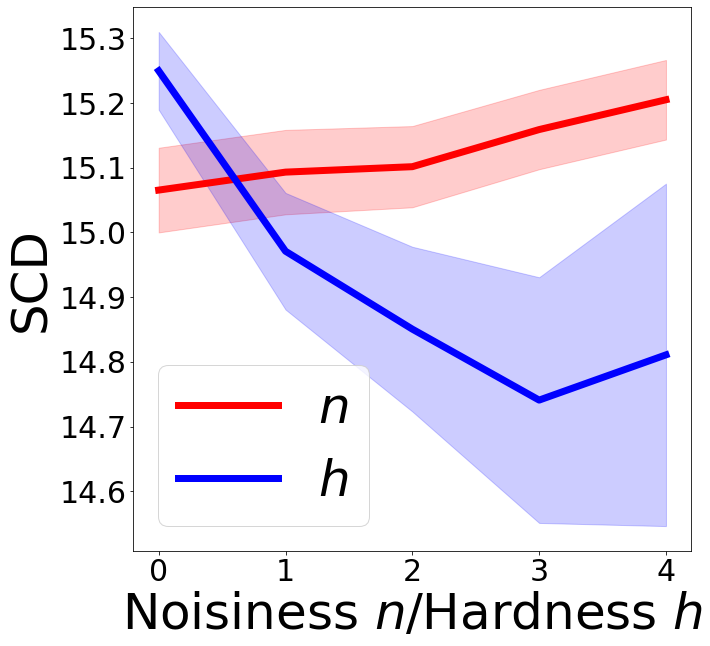}}\quad%
	\subfloat[Hardness via Diversification]{\includegraphics[width=.3\linewidth]{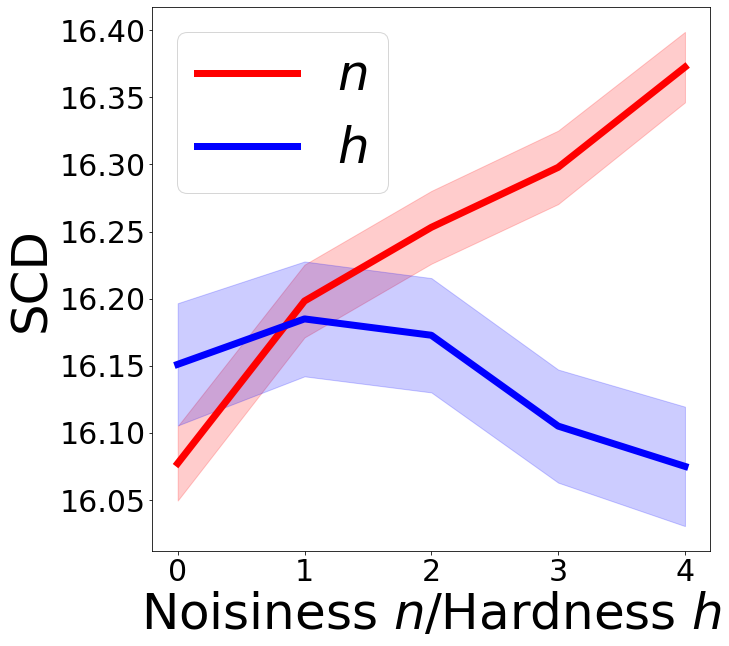}}\quad%
	\subfloat[Hardness via Closeness to the Decision Boundary]{\includegraphics[width=.3\linewidth]{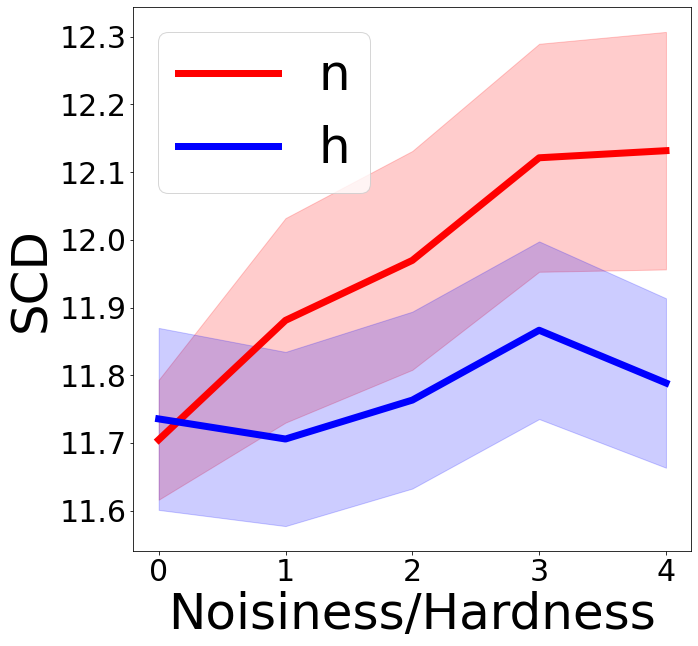}}%
	\caption{ACD (top) and SCD (bottom) applied to the transformed datasets with noisiness level $n$ and hardness level $h$. We observe that SCD increases with $n$, but not with $h$, and is thus non-monotonic in $n$ and $h$. In contrast, ACD is monotonic in $n$ and $h$. This is also observed for other metrics such as loss, confidence, AUL, AUM, and JSD in Figure~\ref{fig:ch5_metrics}. We, therefore, propose SCD as a promising metric to be used in data partitioning for removing noisy samples while retaining hard samples. We quantitatively compare these two metrics with also other metrics in terms of their ability to partition datasets into clean and noisy subsets later in this section in Table~\ref{tab:ch5_partition}.
	}\label{fig:ch5_scd}
\end{figure}

Various label-noise detection and data partitioning methods have been proposed in the literature, including $\text{Thres}_{\text{Loss}}$  \citep{huang2019o2u}, $\text{Thres}_{\text{acc over training}}$ \citep{sun2020crssc}, $\text{Thres}_{\text{AUM}}$ \citep{pleiss2020identifying}, $\text{1d-GMM}_{\text{AUL}}$ \citep{pleiss2020detecting}, and $\text{2d-GMM}_{\text{WJSD}-\text{ACD}}$ \citep{zhang2022combating}.
However, as discussed in Section~\ref{sec:datasets}, metrics such as loss and confidence fail to differentiate between hardness $h$ and noisiness $n$, making it challenging to remove only noisy samples without removing hard ones. The same issue applies to other metrics such as accuracy over training, AUL, AUM, JSD, and ACD, as depicted in Figure~\ref{fig:ch5_metrics} provided in the appendix; 
These metrics are monotonic in both $h$ and $n$.  
Consequently, if these metrics are used for data partitioning, the identified noisy subset may mistakenly contain hard samples, reducing the reliability of these methods when dealing simultaneously with noisy and hard samples. 
In contrast, SCD increases with $n$ but does \emph{not} increase with $h$. We can observe this behavior in Figure~\ref{fig:ch5_scd} and compare it with for example the behavior of ACD when hardness and noisiness increase. This behavior makes SCD a particularly promising metric for removing noisy samples while preserving hard samples. \\
While SCD is a promising metric for distinguishing hard samples from noisy ones, it is beneficial to pair it with another metric that can detect easy samples from the hard and noisy ones. In our experiments, we have observed that accuracy over training is an effective metric for this purpose. It can accurately identify the easy samples in the dataset while mixing the hard and noisy samples together.
Since these two metrics - SCD and accuracy over training - are complementary, we use a 2d-GMM with them as its dimensions to estimate all three subsets - $S_e$, $S_h$, and $S_n$. \\
Table~\ref{tab:ch5_partition} presents the results of different partitioning methods on all three synthetic hard datasets, including the filtered dataset size $|{\tilde{S}}_c|$, correct label percentage of ${\tilde{S}}_c$, $\text{Precision}_{\text{n}}$, $\text{Recall}_{\text{n}}$, and $\text{Recall}_{\text{h}}$. Several interesting observations emerge from the results. Firstly, some methods, such as $\text{Thres}_{\text{AUM}}$, have high $\text{Recall}_{\text{n}}$ but low $\text{Recall}_{\text{h}}$. These methods struggle to differentiate hard samples from noisy ones, and thus discard most of the hard samples. Secondly, some methods, such as $\text{2d-GMM}_{\text{WJSD}-\text{ACD}}$, have high $\text{Recall}_{\text{h}}$ but low $\text{Recall}_{\text{n}}$. Although they preserve hard samples, they also retain noisy ones. Thirdly, these observations vary depending on the type of hardness, making the conclusions less robust. However, we consistently observe that our proposed method, $\text{2d-GMM}_{\text{acc}-\text{SCD}}$, performs the best overall and robustly across all three datasets, with high $\text{Recall}_{\text{n}}$ and $\text{Recall}_{\text{h}}$. Such robustness is particularly crucial in practice since samples can be hard to learn due to any combination of the hardness types, making a reliable metric even more necessary.
\begin{table*}[t]
	\centering
	\small
	\subfloat[Hardness via imbalance dataset]{
		\begin{tabular}{|c|c|c|c|c|c|}
			\hline
			\multirow{2}{*}{{Method}}  &  \multirow{2}{*}{$|{\tilde{S}}_c|$} & Correct Label  & $\text{Precision}_{\text{n}}$ & $\text{Recall}_{\text{n}}$ 
			& $\text{Recall}_{\text{h}}$ \\ 
			& & $\%$ of ${\tilde{S}}_c$ &  $\frac{|{\tilde{S}}_n \cap {{S}}_n|}{|{\tilde{S}}_n|}$ & $\frac{|{\tilde{S}}_n \cap {{S}}_n|}{|{S}_n|}$ &  $\frac{|{\tilde{S}}_c \cap {{S}}_h|}{|{S}_h|}$  \\
			\hline \hline
			Original dataset& $31k$ & $0.80$ & $\texttt{NA}$ & $\texttt{NA}$ & $\texttt{NA}$ \\
			\hline
			$\text{Thres}_{\text{Loss}}$  \citep{huang2019o2u, lu2022selc} & $15.5k$ & $0.88$ & $0.27$ & $0.69$ & $0.03$ \\
			$\text{Thres}_{\text{acc over training}}$ \citep{sun2020crssc} & $15.7k$ & $\mathbf{0.95}$ & $0.35$ & $\mathbf{0.88}$ &  $0.003$ \\
			$\text{Thres}_{\text{AUM}}$ \citep{pleiss2020identifying} & $15.5k$ & $\mathbf{0.95}$ & $0.34$ & $0.86$ &  $0.001$ \\
			\hline
			$\text{1d-GMM}_{\text{Loss}}$ \citep{li2020dividemix} & $29.1k$ & $0.81$ & $0.31$ & $0.10$ & $0.57$ \\
			$\text{1d-GMM}_{\text{AUL}}$ \citep{pleiss2020detecting} & $25.6k$ & $0.88$ & $\mathbf{0.59}$ & $0.52$ & $0.12$ \\
			\hline 
			$\text{2d-GMM}_{\text{WJSD}-\text{ACD}}$ \citep{zhang2022combating} & $18.4k$ & $0.77$ & $0.14$ & $0.30$ & $\mathbf{0.73}$ \\
			$\text{2d-GMM}_{\text{acc}-\text{SCD}}$ (Ours) & $25.7k$ & \underline{$0.83$} & \underline{$0.44$} & \underline{$0.46$} & \underline{$0.61$} \\ 
			\hline
		\end{tabular}
	}\\
	\subfloat[Hardness via diversification dataset]{
		\begin{tabular}{|c|c|c|c|c|c|}
			\hline
			Original dataset& $80k$ & \quad \quad $0.80$ \quad \quad \quad & \quad\quad  $\texttt{NA}$ \quad \quad  & \quad$\texttt{NA}$\quad \quad  & \quad  $\texttt{NA}$\quad  \quad \\
			\hline
			$\text{Thres}_{\text{Loss}}$  \citep{huang2019o2u, lu2022selc} & $40k$ & $0.89$ & $0.29$ & $0.73$ & $0.38$\\
			$\text{Thres}_{\text{acc over training}}$ \citep{sun2020crssc} & $41k$ & $\mathbf{0.96}$ & $0.36$ & $\mathbf{0.90}$ & $0.41$\\
			$\text{Thres}_{\text{AUM}}$ \citep{pleiss2020identifying} & $40k$ & $\mathbf{0.96}$ & $0.35$ & $\mathbf{0.90}$ & $0.37$ \\
			\hline
			$\text{1d-GMM}_{\text{Loss}}$ \citep{li2020dividemix} & $65.8k$ & $0.84$ & $0.38$ & $0.34$ & $0.66$\\
			$\text{1d-GMM}_{\text{AUL}}$ \citep{pleiss2020detecting} & $63k$ & $0.91$ & $\mathbf{0.58}$ & $0.63$ & $\mathbf{0.69}$ \\
			\hline
			$\text{2d-GMM}_{\text{WJSD}-\text{ACD}}$ \citep{zhang2022combating} & $46.1k$ & $0.80$ & $0.19$ & $0.41$ & ${0.50}$ \\
			$\text{2d-GMM}_{\text{acc}-\text{SCD}}$ (Ours) & $62.2k$ & \underline{$0.86$} & \underline{$0.38$} & \underline{$0.43$} & \underline{$0.68$} \\ 
			\hline
		\end{tabular}
	}\\
	\subfloat[Hardness via closeness to the decision boundary dataset]{
		\begin{tabular}{|c|c|c|c|c|c|}
			\hline
			Original dataset& $27.6k$  & \quad \quad $0.80$ \quad \quad \quad & \quad\quad  $\texttt{NA}$ \quad \quad  & \quad$\texttt{NA}$\quad \quad  & \quad  $\texttt{NA}$\quad \quad  \\
			\hline
			$\text{Thres}_{\text{Loss}}$  \citep{huang2019o2u, lu2022selc} & $13.8k$ & $0.93$ & $0.32$ & $0.82$ & $0.46$\\
			$\text{Thres}_{\text{acc over training}}$ \citep{sun2020crssc} & $15.9k$ & $\mathbf{0.99}$ & $0.46$ & $\mathbf{0.99}$ & ${0.59}$ \\
			$\text{Thres}_{\text{AUM}}$ \citep{pleiss2020identifying} &$13.8k$ & $\mathbf{0.99}$ & $0.39$ & $\mathbf{0.99}$ & $0.51$\\
			\hline
			$\text{1d-GMM}_{\text{Loss}}$ \citep{li2020dividemix} & $26.2k$ & $0.84$ & $\mathbf{0.84}$ & $0.22$ & $\mathbf{0.79}$ \\
			$\text{1d-GMM}_{\text{AUL}}$ \citep{pleiss2020detecting} & $19.4k$ & $\mathbf{0.99}$ & $0.65$ & $0.97$ & $0.70$\\
			\hline 
			$\text{2d-GMM}_{\text{WJSD}-\text{ACD}}$ \citep{zhang2022combating} & $22k$ & $0.85$ & $0.37$ & $0.38$ &  ${0.67}$ \\
			$\text{2d-GMM}_{\text{acc}-\text{SCD}}$ (Ours) & $23.1k$ & \underline{$0.92$} & \underline{$0.81$} & \underline{$0.68$} & \underline{$0.77$} \\ 
			\hline
		\end{tabular}
	}
	\caption{Partitioning results obtained by applying various label noise detection methods on our three synthetic datasets with hard samples. We can observe that, on the one hand, some methods have a good performance on recall only on noisy samples  $\text{Recall}_{\text{n}}$ (for example $\text{Thres}_{\text{acc over training}}$), whereas they have very bad performance in terms of recalling hard samples (low $\text{Recall}_{\text{h}}$). This means that, in an attempt to remove noisy samples from the dataset, they remove almost all hard samples as well. On the other hand, some methods have a good performance in terms of keeping hard samples (high $\text{Recall}_{\text{h}}$) (for example $\text{2d-GMM}_{\text{WJSD}-\text{ACD}}$), but fail to remove the noisy samples from the training dataset, as evidenced by the very low value of $\text{Recall}_{\text{n}}$. Our proposed metric, $\text{2d-GMM}_{\text{acc}-\text{SCD}}$, shows the best overall performance in terms of removing noisy samples while keeping hard samples, as evidenced by the relatively high values of $\text{Recall}_{\text{h}}$ and $\text{Recall}_{\text{n}}$.  The result is consistent in all three settings, unlike some methods that only perform well in one of the hardness types.}
	\label{tab:ch5_partition}
\end{table*}


\subsection{Training on the Filtered Subset}

In this section, we present the results of training models on the estimated clean datasets $\tilde{S}_c$ obtained in the previous section.  Table~\ref{tab:ch5_trainonclean_1} displays the generalization performance of models trained on the estimated clean datasets using each method. The results demonstrate that our proposed method $\text{2d-GMM}_{\text{acc}-\text{SCD}}$ consistently outperforms the other methods in all three hardness types. We observe that the $\text{1d-GMM}_{\text{AUL}}$ method performs slightly better than $\text{2d-GMM}_{\text{acc}-\text{SCD}}$ in the second and third hardness types. However, in the first hardness type, $\text{1d-GMM}_{\text{AUL}}$ performs significantly worse than $\text{2d-GMM}_{\text{acc}-\text{SCD}}$, which suggests that $\text{1d-GMM}_{\text{AUL}}$ is not robust to the hardness type and thus unreliable for practical use.\\
Overall, the table emphasizes the importance of selecting an appropriate metric/approach for data filtration in presence of both noisy and hard samples. Moreover, our experiments highlight the advantage of using the label noise detection method $\text{2d-GMM}_{\text{acc}-\text{SCD}}$ for different hardness types. 

\begin{table}[t]
\centering
	\small
 \begin{adjustbox}{max width=\textwidth}
    \begin{tabular}{|c|c|c|c|}
			\hline
			Method & Hardness via Imbalance & Hardness via Diversification & Hardness via Closeness to the Decision Boundary \\ 
			\hline \hline$\text{Thres}_{\text{Loss}}$  \citep{huang2019o2u, lu2022selc}& $25.83_{\pm 0.06}\%$& $42.65_{\pm 0.07}\%$&$43.30_{\pm 0.33}\%$\\
			$\text{Thres}_{\text{acc over training}}$ \citep{sun2020crssc} &$25.34_{\pm 0.42}\%$ & $44.39_{\pm 0.16}\%$&$49.13_{\pm 0.43}\%$\\
			$\text{Thres}_{\text{AUM}}$ \citep{pleiss2020identifying} & $23.85_{\pm 0.21}\%$& $44.35_{\pm 0.41}\%$&$47.67_{\pm 0.09}\%$\\
			\hline
			$\text{1d-GMM}_{\text{Loss}}$ \citep{li2020dividemix} &$31.08_{\pm 0.13}\%$ & $42.93_{\pm 0.37}\%$&$44.93_{\pm 0.17}\%$\\
			$\text{1d-GMM}_{\text{AUL}}$ \citep{pleiss2020detecting} &$30.27_{\pm 0.34}\%$ & $\mathbf{45.56}_{\pm 0.39}\%$&$\mathbf{50.67}_{\pm 0.26}\%$\\
			\hline 
			$\text{2d-GMM}_{\text{WJSD}-\text{ACD}}$ \citep{zhang2022combating} &$32.11_{\pm 0.15}\%$ &${38.83}_{\pm 0.34}\%$ &$43.23_{\pm 0.21}\%$\\
			$\text{2d-GMM}_{\text{acc}-\text{SCD}}$ (Ours)  & $\mathbf{34.88_{\pm 0.13}}\%$& $\mathbf{45.36}_{\pm 0.33}\%$&$\mathbf{49.80}_{\pm 0.24}\%$\\
			\hline
		\end{tabular}
  \end{adjustbox}
  \caption{DenseNet test accuracy comparison on the estimated clean datasets using different data filtration methods for datasets with different hardness types. Our method $\text{2d-GMM}_{\text{acc}-\text{SCD}}$ consistently performs well in all hardness types, unlike 
		the $\text{1d-GMM}_{\text{AUL}}$ method that performs well only in the second and third hardness types. It is important to choose a method that works well with all hardness types because, in practice, hard samples can arise from any of the three types tested here.}\label{tab:ch5_trainonclean_1} 
\end{table}

\begin{table}[h]
	\centering\small
 \begin{adjustbox}{max width=\textwidth}
 \begin{tabular}{|c|c|c|c|c|c|c|}
            \hline
            Dataset & \multicolumn{3}{c|}{Animal-10N} & \multicolumn{3}{c|}{CIFAR-10N} \\
		\hline
		\backslashbox{Method}{Performance Metric} & Test Accuracy $\%$ & Test Loss & Estimated LNL & Test Accuracy $\%$ & Test Loss & Estimated LNL \\ 
		\hline \hline
		$\text{Thres}_{\text{Loss}}$ \citep{huang2019o2u, lu2022selc} & $80.32_{\pm 0.38}$ & $0.62_{\pm 0.01}$ & $50\%$& $85.45_{\pm 0.22}$ & $0.64_{\pm 0.01}$ & $50\%$\\
		$\text{Thres}_{\text{acc over training}}$ \citep{sun2020crssc} & $83.03_{\pm 0.11}$ & $0.52_{\pm 0.01}$ & $2.48\%$& $83.61_{\pm 0.17}$ & $0.78_{\pm 0.01}$ & $46\%$\\
		$\text{Thres}_{\text{AUM}}$ \citep{pleiss2020identifying} & $76.30_{\pm 0.16}$ & $0.90_{\pm 0.04}$ & $50\%$& $81.89_{\pm 0.18}$ & $0.83_{\pm 0.01}$ & $50\%$\\
		\hline
		$\text{1d-GMM}_{\text{Loss}}$ \citep{li2020dividemix} &$83.05_{\pm 0.25}$ & $0.53_{\pm 0.01}$ & $2.82\%$& $85.87_{\pm 0.30}$ & $0.62_{\pm 0.00}$ & $46\%$\\
		$\text{1d-GMM}_{\text{AUL}}$ \citep{pleiss2020detecting} &${82.80}_{\pm 0.28}$ & ${0.53}_{\pm 0.01}$ & $2.11\%$& $86.21_{\pm 0.06}$ & $0.66_{\pm 0.01}$ & $30\%$\\
		\hline 
		$\text{2d-GMM}_{\text{WJSD}-\text{ACD}}$ \citep{zhang2022combating} &$83.06_{\pm 0.26}$ &${0.54_{\pm 0.01}}$ & $1.03\%$& $89.44_{\pm 0.22}$ & $0.35_{\pm 0.01}$& $17\%$\\
		$\text{2d-GMM}_{\text{acc}-\text{SCD}}$ (Ours)  & $\mathbf{83.11}_{\pm 0.10}$& $\mathbf{0.51}_{\pm 0.01}$ & $\mathbf{3.3}\%$ & $\mathbf{89.90}_{\pm 0.21}$ & $\mathbf{0.31}_{\pm 0.00}$ & $\mathbf{10}\%$\\
		\hline
	\end{tabular}
 \end{adjustbox}
	\caption{Performance comparison of models trained on the estimated clean datasets using different data filtration methods for the Animal-10N and CIFAR-10N datasets. The estimated label noise level (LNL) of Animal-10N and CIFAR-10N are around $8\%$ and $9.01\%$, respectively. We observe that our method provides the best LNL estimate. The LNL estimate of each method is computed using $1-\abs{\tilde{S}_c}/\abs{S}$. Moreover, our method $\text{2d-GMM}_{\text{acc}-\text{SCD}}$ provides the cleanest dataset as evidenced by the large margin in the generalization performance improvement; both test accuracy and test loss are the best for $\text{2d-GMM}_{\text{acc}-\text{SCD}}$.}\label{tab:ch5_trainonly_animal10n}\label{tab:ch5_trainonly_cifar10n}
\end{table}

\textbf{Experiments on datasets with real-world label noise} 
To further evaluate and compare the performance of each data filtration method, we apply them to datasets with real-world label noise. Table~\ref{tab:ch5_trainonly_animal10n} displays the results of each method for partitioning and data filtration, for models trained on the Animal-10N\footnote{\url{https://dm.kaist.ac.kr/datasets/animal-10n/}} and CIFAR-10N \citep{wei2022learning} datasets. Both datasets have real-world label noise and unknown easy-hard-noisy subsets. Since we lack knowledge of which samples are hard or incorrectly labeled, we cannot compute $\text{Recall}_{\text{h}}$ or $\text{Recall}_{\text{n}}$, as we did for our created synthetic datasets in Table~\ref{tab:ch5_partition}. Nonetheless, we can apply each partitioning/label-noise detection method to these datasets, partition them into an estimated clean subset $\tilde{S}_c$ and an estimated noisy subset $\tilde{S}_n$, and train models on $\tilde{S}_c$. The generalization performance of models trained on $\tilde{S}_c$ obtained from each method is an indication of the quality of the estimated clean subsets $\tilde{S}_c$. Our results in Table~\ref{tab:ch5_trainonly_animal10n} demonstrate that our proposed method, $\text{2d-GMM}_{\text{acc}-\text{SCD}}$, outperforms all other methods by a significant margin in terms of test performance and in terms of estimating the label noise level of the given dataset.

\subsection{Semi-supervised Learning on the Filtered Subsets}
After the data partitioning step to partition the dataset into an estimated clean subset $\tilde{S}_c$ and an estimated noisy subset $\tilde{S}_n$, we can then apply a semi-supervised learning algorithm and use $\tilde{S}_c$ as the labeled set and $\tilde{S}_n$ as the unlabeled set (by discarding the noisy labels in $\tilde{S}_n$). We use the Flex-Match semi-supervised learning algorithm \citep{zhang2021flexmatch}, which is shown to have good performance. The results are shown in Table~\ref{tab:flexmatch}, for the two real-world datasets Animal-10N and CIFAR-10N. We can observe the significant performance improvement brought by using $\text{2d-GMM}_{\text{acc}-\text{SCD}}$ as a data partitioning method, compared to other partitioning methods. This once again indicates that our data partitioning method is able to remove incorrectly labeled samples that do not help generalization, and to further retain hard samples which help generalization.\\
Note that semi-supervised learning algorithms are well-suited for settings where there is a large set of unlabeled data and a small set of labeled data. However, in our settings, where we use different data partitioning methods on the Animal-10N and CIFAR-10N datasets which are datasets with low label noise levels, the estimated noisy subset is not large. Hence, algorithms such as Flex-Match require a large number of epochs in order to get to convergence. This was computationally expensive and hence we present results of networks that are stopped at a training loss value of $0.45$. This is the reason we are observing a relatively lower performance when using a semi-supervised learning approach compared to training only on the filtered labeled dataset (reported in Table~\ref{tab:ch5_trainonly_animal10n}).

\begin{table}[h]
	\centering\small
 \begin{tabular}{|c|c|c|}\hline
		\backslashbox{Method}{Dataset} & Animal-10N & CIFAR-10N \\ 
		\hline \hline
		$\text{Thres}_{\text{Loss}}$ \citep{huang2019o2u, lu2022selc} & $\mathbf{77.84}_{\pm 0.50}$ & $63.85_{\pm 2.25}$ \\
		$\text{Thres}_{\text{acc over training}}$ \citep{sun2020crssc} & $71.60_{\pm 0.54}$ & $69.44_{\pm 0.18}$ \\
		$\text{Thres}_{\text{AUM}}$ \citep{pleiss2020identifying} & $72.40_{\pm 0.39}$ & $61.32_{\pm 0.57}$ \\
		\hline
		$\text{1d-GMM}_{\text{Loss}}$ \citep{li2020dividemix} & $72.15_{\pm 1.81}$ & $65.67_{\pm 0.84}$\\
		$\text{1d-GMM}_{\text{AUL}}$ \citep{pleiss2020detecting} & $70.11_{\pm 1.55}$ & $66.65_{\pm 0.94}$ \\
		\hline 
		$\text{2d-GMM}_{\text{WJSD}-\text{ACD}}$ \citep{zhang2022combating} & $69.88_{\pm 0.67}$ & $76.43_{\pm 0.80}$\\
		$\text{2d-GMM}_{\text{acc}-\text{SCD}}$ (Ours)  & $\mathbf{77.43}_{\pm 0.46}$ & $\mathbf{80.45}_{\pm 0.20}$ \\
		\hline
	\end{tabular}
	\caption{Test accuracy percentage comparison for models trained using Flex-Match algorithm on the partitioned datasets found from each method. The estimated clean subset $\tilde{S}_c$ and noisy subset $\tilde{S}_n$ are used as the labeled and unlabeled sets, respectively. We can observe that $\text{2d-GMM}_{\text{acc}-\text{SCD}}$ achieves the best dataset quality as evidenced by the high test accuracy of the trained models in both Animal-10N and CIFAR-10N datasets.
	}
	\label{tab:flexmatch} 
\end{table}

\section{Discussion}

\textbf{Robustness to Hardness Type}
In this study, we aim to explore different approaches for manipulating the hardness of samples and investigate their impact on label noise detection methods. We simulate three hardness types and compared the performance of various methods in identifying and removing noisy samples while retaining hard ones. It is important to highlight that sample hardness level $h$ is a relative concept and can only be evaluated when comparing samples within a given dataset. This is different from the noisiness level $n$, which is an absolute measure. Our simulations showed that the three hardness types that we test are sufficiently distinct. Interestingly, we find that some methods perform better in distinguishing certain types of hard samples from noisy ones, while not performing well in distinguishing other hard samples. However, the 2D-GMM on top of accuracy over training and SCD, demonstrated the most robust performance across all hardness types. The reason for the rather significant superior performance of this method in real-world datasets stems from its ability to effectively detect label noise and hard samples caused by various underlying reasons, which are often unpredictable in real-world datasets. When proposing a label noise detection method, it is essential to consider the unpredictability of hard samples in real-world datasets.\\
\begin{figure}[t]
	\centering
	\subfloat[Easy/Hard/Noisy Clusters]{\includegraphics[width=.4\linewidth]{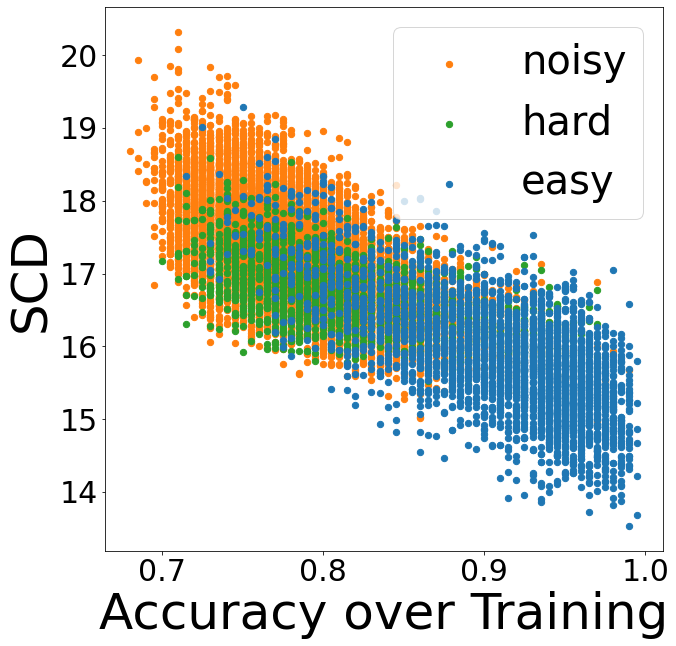}}\quad
	\subfloat[Results of a GMM with 2 Clusters]{\includegraphics[width=.4\linewidth]{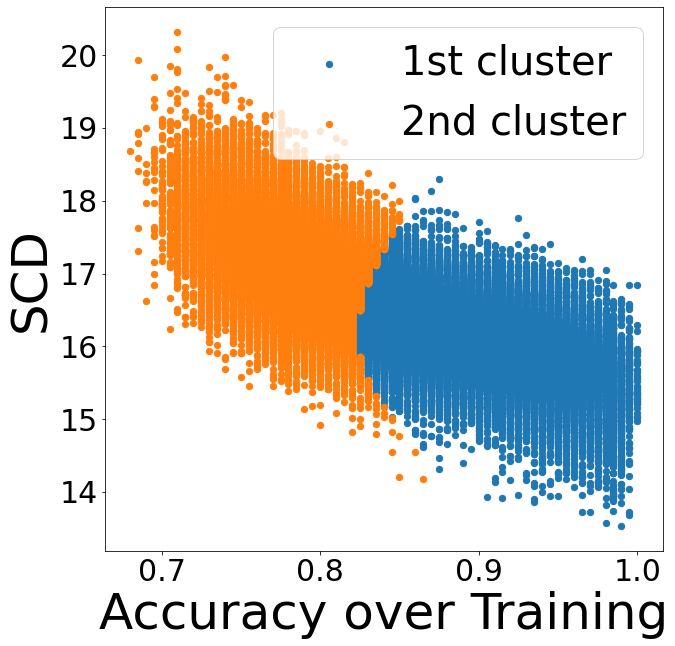}}\\
	\subfloat[Results of a GMM with 3 Clusters]{\includegraphics[width=.4\linewidth]{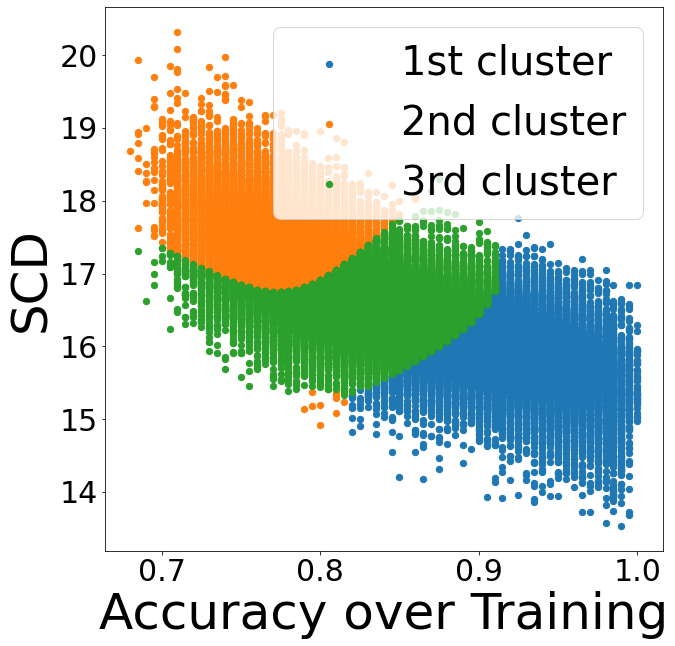}}
	\caption{(a) Easy-hard-noisy clusters for samples of the hardness via diversification dataset with their associated static centroid distance (SCD) and accuracy over training values. We can observe that noisy samples are located on the top left of the figure, followed by the hard samples in the middle parts and the easy samples on the other end, i.e., the bottom right. Easy samples have a high accuracy over training and a low SCD. Depicting such figures requires full ground-truth access to the available dataset and knowledge about label noise and hardness levels of samples. However, this information is not possible in practice, hence in the sub-figures (b) and (c) we evaluate the results of Gaussian mixture models to recover these clusters without any knowledge about samples. (b) Clustering results of $\text{2d-GMM}_{\text{acc}-\text{SCD}}$ with $2$ number of clusters. This clustering does not require the ground-truth access of samples and only requires the computation of SCD and accuracy over training. If we compare the clusters with actual easy-hard-noisy partitions in sub-figure (a) we can observe that most of the hard and noisy samples are mixed in the second cluster. (c) Clustering results of $\text{2d-GMM}_{\text{acc}-\text{SCD}}$ with $3$ number of clusters. In contrast to clustering with $2$ clusters, we can observe that clustering with $3$ cluster is much better at including as many noisy samples as possible in the second cluster while not including hard samples. Hence, throughout our study, when referring to our label noise detection method $\text{2d-GMM}_{\text{acc}-\text{SCD}}$, we apply $3$ clusters and use the top left cluster as the detected noisy subset.
	}\label{fig:ch5_clusters}
\end{figure}
\textbf{Number of Clusters in GMM}
We observe that in our synthetic datasets, the location of easy $S_e$, hard $S_h$, and noisy $S_n$ samples in the 2D spectrum with accuracy over training and SCD followed a certain pattern. We illustrate this pattern in Figure~\ref{fig:ch5_clusters} for the hardness via diversification dataset. Furthermore, we find that this pattern is consistent across different hardness types. To further investigate this observation, we compare GMM models with two and three clusters and analyze the resulting clusters. Our observation is that when we use two clusters, many hard and noisy samples are clustered together, resulting in poor detection performance. However, when we use three clusters, the detection performance improves significantly, and the resulting clusters are much more coherent with the actual clusters. This observation led us to use a 3-cluster GMM in our proposed method, $\text{2d-GMM}_{\text{acc}-\text{SCD}}$, which produced more accurate results.
It is important to note that such investigation is only possible through our controlled experiments on synthetic datasets because we know the exact partitions of the datasets into easy, hard, and noisy samples. Nevertheless, our findings provide valuable insights into the design of label noise detection methods even when used with real-world datasets.

\textbf{Conclusion} 
We propose an empirical approach to investigate hard samples in an image classification setting. To this end, we create synthetic datasets that enable us to study hard and noisy samples in a controlled environment. Through our investigation, we make several interesting observations, including the importance of analyzing feature layer vectors of neural networks to distinguish between hard and noisy samples. Furthermore, we propose a label noise detection method that outperforms other existing methods in terms of both removing noisy samples and retaining hard ones. Our method can be applied to filter datasets with label noise, leading to better generalization performance when training models on the filtered set. Importantly, our label noise detection method is of quite general use and can be used in combination with any other method designed to deal with label noise and/or hard samples. Although our synthetic datasets were tailored toward image classification tasks, our conclusions could be applied to other applications. Moreover, our data filtration method could be combined with semi-supervised learning methods, such as FixMatch and FlexMatch, to further improve the model's generalization performance by making use of the discarded subset of the data. Overall, our study provides valuable insights into the design and optimization of label noise detection methods and their applications in improving the performance of machine learning models in real-world settings.

\clearpage

\bibliographystyle{abbrvnat}
\bibliography{ms}

\appendix
\section{Experimental Setup}\label{sec:ch5_expsetup}

\textbf{Generating Noisy-labeled Datasets} We modify original datasets similarly to \cite{chatterjee2020coherent}; for a fraction of samples denoted by the label noise level (LNL), we replace the labels with independent random variables drawn uniformly from other classes with the same noisiness level $n$. 

For each hardness type, there are two sets of experiments: (i) experiments done to detect clean-noisy subsets, (ii) experiments done to train on filtered sets. We provide details for each of these experiments below.

\textbf{Hardness via Imbalance Experiments} 
(i) The models are trained for $200$ epochs on the cross-entropy objective function with batch size $100$, using SGD with learning rate $0.001$, weight decay $5\cdot10^{-4}$, and momentum $0.9$. The maximum label noise is set as $40\%$. The neural network architecture is MobileNetV2. (ii) The models are trained for $50$ epochs on the cross-entropy objective function with batch size $100$, using SGD with learning rate $0.001$, weight decay $5\cdot10^{-4}$, and momentum $0.9$. The neural network architecture is DenseNet121, pre-trained on ImageNet dataset.

\textbf{Hardness via Diversification Experiments} 
(i) The models are trained for $200$ epochs on the cross-entropy objective function with batch size $100$, using SGD with learning rate $0.001$, weight decay $5\cdot10^{-4}$, and momentum $0.9$. The maximum label noise is set as $40\%$. The neural network architecture is MobileNetV2. (ii) The models are trained for $50$ epochs on the cross-entropy objective function with batch size $100$, using SGD with learning rate $0.001$, weight decay $5\cdot10^{-4}$, and momentum $0.9$. The neural network architecture is DenseNet121, pre-trained on ImageNet dataset.

\textbf{Hardness via Closeness to the Decision Boundary Experiments} 
(i) The models are trained for $50$ epochs on the cross-entropy objective function with batch size $100$, using SGD with learning rate $0.001$, weight decay $5\cdot10^{-4}$, and momentum $0.9$. The maximum label noise is set as $40\%$. The neural network architecture is DesneNet121, pre-trained on ImageNet. (ii) The models are trained for $50$ epochs on the cross-entropy objective function with batch size $100$, using SGD with learning rate $0.001$, weight decay $5\cdot10^{-4}$, and momentum $0.9$. The neural network architecture is DenseNet121, pre-trained on ImageNet dataset.

\textbf{CIFAR-10N Experiments} 
The models are trained for $50$ epochs on the cross-entropy objective function with batch size $64$, using SGD with learning rate $0.05$, weight decay $5\cdot10^{-4}$, and momentum $0.9$. The neural network architecture is MobileNetV2.

\textbf{Animal-10N Experiments} 
The models are trained for $40$ epochs on the cross-entropy objective function with batch size $64$, using SGD with learning rate $0.001$, weight decay $5\cdot10^{-4}$, and momentum $0.9$. The neural network architecture is AlexNet.

\section{Comparison to Other Metrics}\label{sec:ch5_metrics}
\begin{figure}[t]
	\centering
	\subfloat{\includegraphics[width=.25\linewidth]{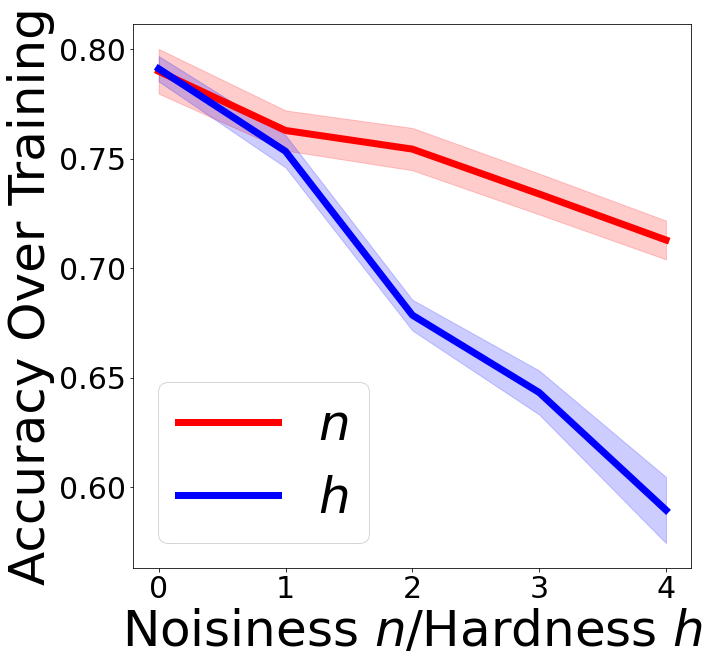}}\quad%
	\subfloat{\includegraphics[width=.25\linewidth]{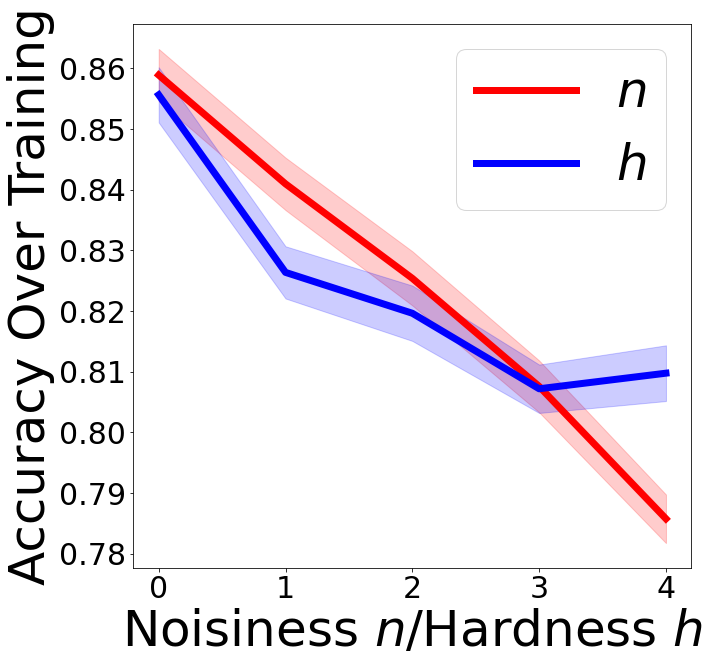}}\quad%
	\subfloat{\includegraphics[width=.25\linewidth]{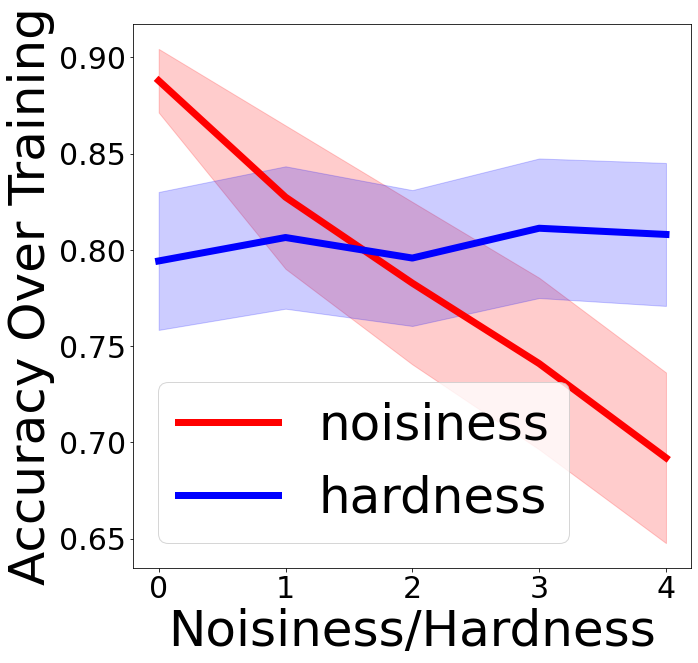}}\\
	\subfloat{\includegraphics[width=.25\linewidth]{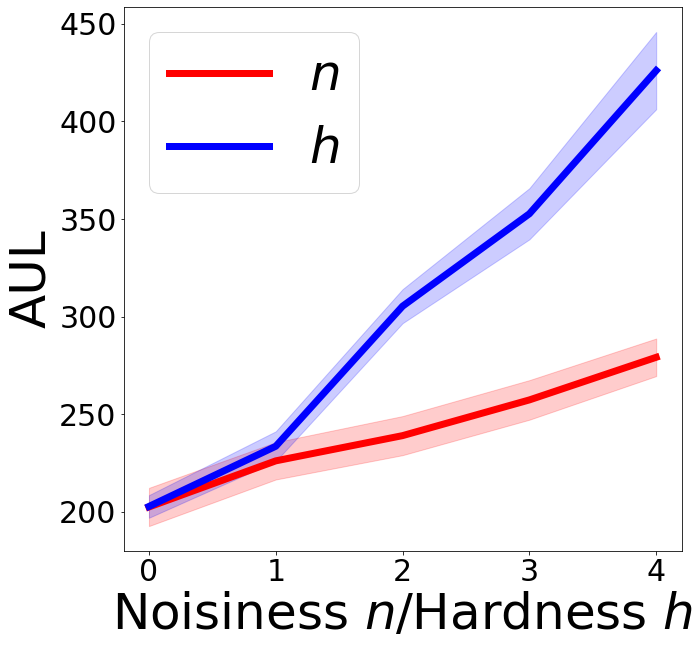}}\quad%
	\subfloat{\includegraphics[width=.25\linewidth]{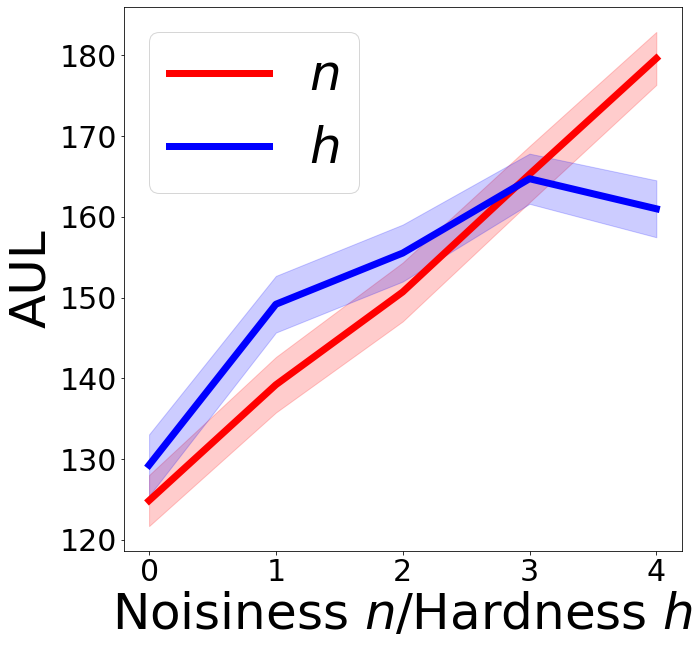}}\quad%
	\subfloat{\includegraphics[width=.25\linewidth]{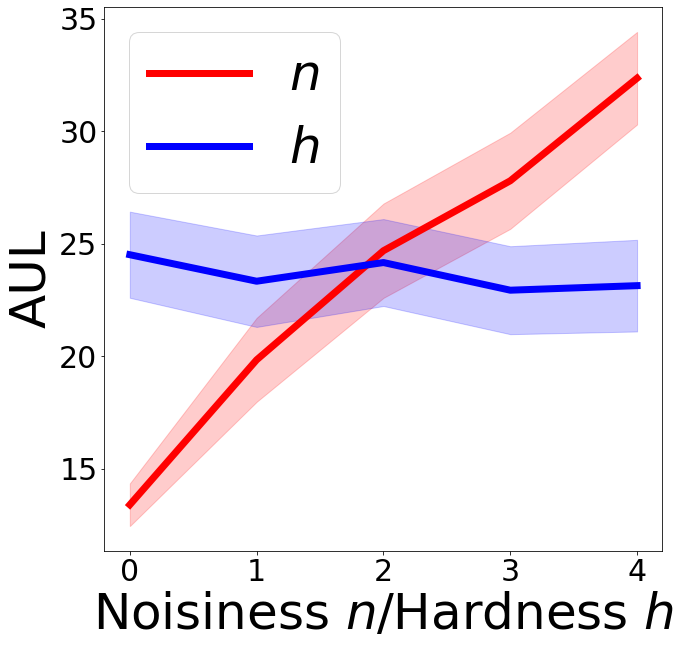}}\\
	\subfloat{\includegraphics[width=.25\linewidth]{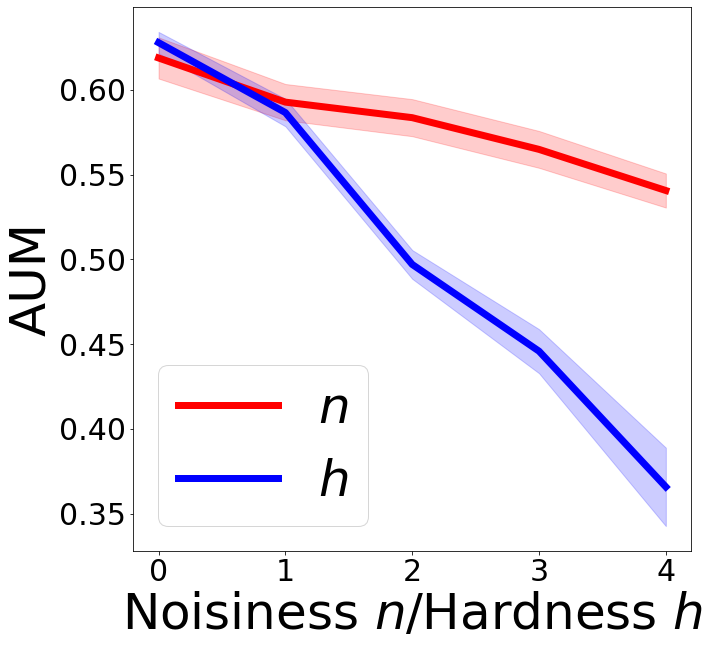}}\quad%
	\subfloat{\includegraphics[width=.25\linewidth]{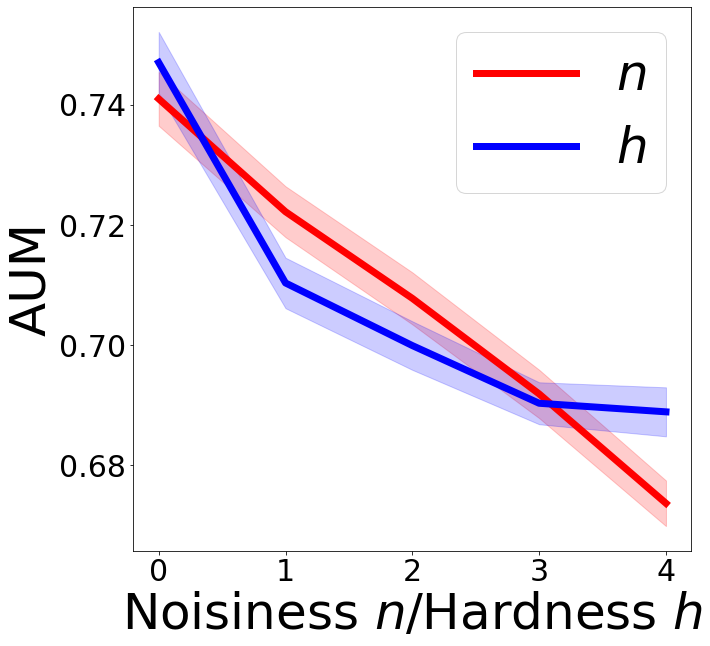}}\quad%
	\subfloat{\includegraphics[width=.25\linewidth]{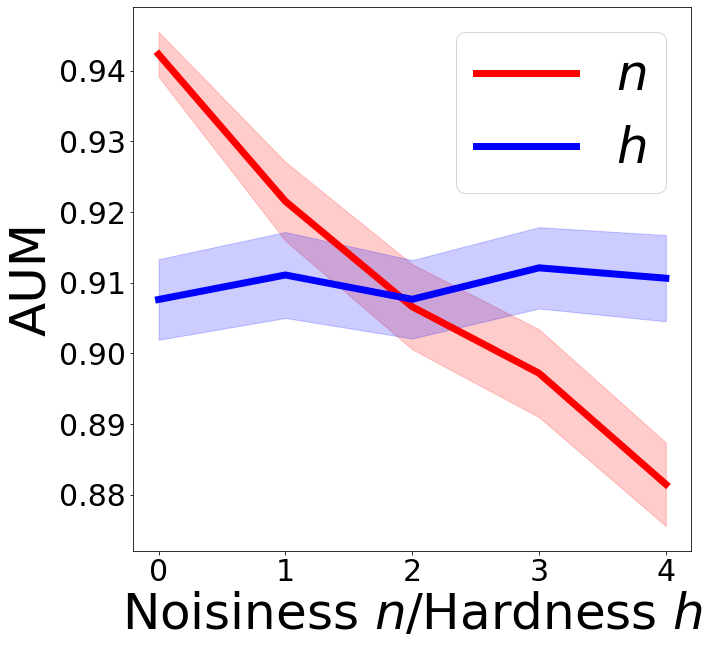}}\\
	\subfloat{\includegraphics[width=.25\linewidth]{acd_ex1.png}}\quad%
	\subfloat{\includegraphics[width=.25\linewidth]{acd_ex2.png}}\quad%
	\subfloat{\includegraphics[width=.25\linewidth]{acd_ex3.png}}\\
	\subfloat[Hardness via Imbalance]{\includegraphics[width=.25\linewidth]{scd_ex1.png}}\quad%
	\subfloat[Hardness via Diversification]{\includegraphics[width=.25\linewidth]{scd_ex2.png}}\quad%
	\subfloat[Hardness via Closeness to the Decision Boundary]{\includegraphics[width=.25\linewidth]{scd_ex3.png}}%
		\caption{Different metrics applied to the transformed datasets with noisiness level $n$ and hardness level $h$. Left, middle and right figures correspond to hardness types imbalanced, diversification, and closeness to the decision boundary, respectively. We observe that all the metrics have at least one type of hardness $h$ that with its increase, the metric shows the same behavior as with increase of noisiness $n$. In contrast, SCD is the only metric that shows different behaviors with increases in $h$ and $n$.
	}\label{fig:ch5_metrics}
\end{figure}

Figure~\ref{fig:ch5_metrics} shows a comparison of different metrics and how they behave concerning changes in the hardness level $h$ and noisiness level $n$. From the graph, we can see that all metrics either increase or decrease as the value of $n$ increases. However, we also observe that the same increase or decrease happens when at least one type of hardness $h$ increases.
In contrast, the SCD metric stands out as it does not show the same trend towards an increase in any type of hardness level $h$ as the value of $n$ increases.

\textbf{Detailed comparison between $\text{2d-GMM}_{\text{WJSD}-\text{ACD}}$ and $\text{2d-GMM}_{\text{acc}-\text{SCD}}$}
Table~\ref{tab:ablation} shows a transition from the partitioning method introduced in \cite{zhang2022combating}, $\text{2d-GMM}_{\text{WJSD}-\text{ACD}}$, to our proposed method, $\text{2d-GMM}_{\text{acc}-\text{SCD}}$. A few findings can be concluded from this table. Computing the ACD metric in the middle of the training improves $\text{Recall}_{\text{n}}$. In addition, using a Euclidean norm instead of cosine, and using the static centroid instead of adaptive, improves $\text{Recall}_{\text{h}}$. On one hand, the most improvement on $\text{Recall}_{\text{h}}$ is achieved by transitioning from two-cluster GMM to three-cluster GMM. On the other hand, the most improvement on $\text{Recall}_{\text{n}}$ is achieved by transitioning from using WJSD to using accuracy over training. Such transition may degrade however $\text{Recall}_{\text{h}}$. Hence, the best overall performance is achieved if we use the combination of all these transitions, i.e., computing feature distance in the middle of the training, using Euclidean norm, taking the static centroid distance, using three clusters in the GMM, and using the accuracy over training. Such transitions result in the $\text{2d-GMM}_{\text{acc}-\text{SCD}}$ method, which is what we propose as our data partitioning method.

\begin{table*}[t]
	\centering
	\small
		\begin{tabular}{|c|c|c|c|c|c|}
			\hline
			\multirow{2}{*}{{Method}}  &  \multirow{2}{*}{$|{\tilde{S}}_c|$} & Correct Label  & $\text{Precision}_{\text{n}}$ & $\text{Recall}_{\text{n}}$ 
			& $\text{Recall}_{\text{h}}$ \\ 
			& & $\%$ of ${\tilde{S}}_c$ &  $\frac{|{\tilde{S}}_n \cap {{S}}_n|}{|{\tilde{S}}_n|}$ & $\frac{|{\tilde{S}}_n \cap {{S}}_n|}{|{S}_n|}$ &  $\frac{|{\tilde{S}}_c \cap {{S}}_h|}{|{S}_h|}$  \\
			\hline \hline
			Original dataset& $31k$ & $0.80$ & $\texttt{NA}$ & $\texttt{NA}$ & $\texttt{NA}$ \\
			\hline
			$\text{2d-GMM}_{\text{WJSD}-\text{ACD}}$ \citep{zhang2022combating} & $23.5k$ & $0.842$ & $0.325$ & $0.394$ & ${0.191}$ \\
			\hline
                $\text{2d-GMM}_{\text{WJSD}-\text{ACD}_{\text{mid}}}$ & $23.5k$ & $0.845$ & $0.332$ & $0.408$ & ${0.196}$ \\
                $\text{2d-GMM}_{\text{WJSD}-\text{ACD}_{\text{mid}-\text{norm}}}$ & $23.6k$ & $0.842$ & $0.325$ & $0.391$ & ${0.198}$ \\
                $\text{2d-GMM}_{\text{WJSD}-\text{ACD}_{\text{mid}-\text{static}}}$ & $23.6k$ & $0.842$ & $0.325$ & $0.392$ & ${0.201}$ \\
                $\text{2d-GMM-3clusters}_{\text{WJSD}-\text{ACD}_{\text{mid}}}$ & $28.6k$ & $0.814$ & $0.344$ & $0.132$ & ${0.541}$ \\
                $\text{2d-GMM-3clusters}_{\text{WJSD}-\text{ACD}_{\text{mid}-\text{norm}}}$ & $28k$ & $0.819$ & $0.351$ & $0.172$ & ${0.485}$ \\
                $\text{2d-GMM-3clusters}_{\text{WJSD}-\text{ACD}_{\text{mid}-\text{static}}}$ & $28.7k$ & $0.813$ & $0.337$ & $0.127$ & ${0.542}$ \\
                $\text{2d-GMM-3clusters}_{\text{acc}-\text{ACD}}$ & $21.7k$ & $0.912$ & $0.452$ & $0.688$ & ${0.009}$ \\
                \hline
                $\text{2d-GMM}_{\text{acc}-\text{SCD}}$ (Ours) & $24.1k$ & \underline{$0.892$} & \underline{$0.513$} & \underline{$0.576$} & \underline{$0.473$} \\ 
			\hline
		\end{tabular}
 \caption{Data partitioning results on the hardness via imbalance dataset comparing different methods that range from using the proposed method in \citep{zhang2022combating} to using our proposed method, $\text{2d-GMM}_{\text{acc}-\text{SCD}}$. We can observe that each modification that is made from the method $\text{2d-GMM}_{\text{WJSD}-\text{ACD}}$ either improves $\text{Recall}_{\text{n}}$ or $\text{Recall}_{\text{h}}$. Overall, we can observe that method $\text{2d-GMM}_{\text{acc}-\text{SCD}}$ outperforms method $\text{2d-GMM}_{\text{WJSD}-\text{ACD}}$ in terms of removing incorrectly labeled samples while retaining hard samples.}
	\label{tab:ablation}
\end{table*}

\end{document}